\documentclass[11pt, a4paper, twocolumn, confidential, copyright, goog]{google}

\usepackage[authoryear, sort&compress, round]{natbib}
\renewcommand{\today}{2026-10-5}
\setboolean{confidential}{false}
\uselogo{} 

\usepackage{booktabs}
\usepackage{tabularx}
\usepackage{array}
\usepackage{multirow}
\usepackage{makecell}
\usepackage{placeins}
\usepackage{xurl}
\usepackage{flushend}
\AtBeginDocument{\raggedend}
\usepackage{amsmath,amssymb}
\usepackage{tikz}
\usetikzlibrary{arrows.meta,positioning,fit,calc,shapes.geometric}

\usepackage{microtype}
\usepackage{algorithm}
\usepackage{algorithmic}
\usepackage{newfloat}
\usepackage{listings}
\DeclareCaptionStyle{ruled}{labelfont=normalfont,labelsep=colon,strut=off} % DO NOT CHANGE THIS
\floatstyle{ruled}
\newfloat{listing}{tb}{lst}{}
\floatname{listing}{Listing}

\extrafloats{100}

\makeatletter
\newcommand{\webrider@autorefname}[1]{%
  \def\webrider@prefix{#1}%
  \def\webrider@tab{tab}%
  \def\webrider@fig{fig}%
  \def\webrider@sec{sec}%
  \def\webrider@app{app}%
  \def\webrider@lst{lst}%
  \def\webrider@alg{alg}%
  \ifx\webrider@prefix\webrider@tab Table%
  \else\ifx\webrider@prefix\webrider@fig Figure%
  \else\ifx\webrider@prefix\webrider@sec Section%
  \else\ifx\webrider@prefix\webrider@app Appendix%
  \else\ifx\webrider@prefix\webrider@lst Listing%
  \else\ifx\webrider@prefix\webrider@alg Algorithm%
  \else Reference%
  \fi\fi\fi\fi\fi\fi
}
\def\webrider@autoref#1:#2\@nil{\webrider@autorefname{#1}}
\providecommand{\autoref}[1]{\webrider@autoref#1:\@nil~\ref{#1}}
\makeatother

\newcommand{\cgs}{\ensuremath{\mathrm{CGS}}}

\newcommand{\NumBaseTasks}{768}
\newcommand{\NumSites}{42}
\newcommand{\NumDomains}{12}
\newcommand{\NumPersonaPolicies}{15}
\newcommand{\NumContracts}{4,096}
\newcommand{\NumTrainContracts}{2,560}
\newcommand{\NumDevContracts}{512}
\newcommand{\NumTestContracts}{1,024}

\newcommand{\NumTaskAuditPass}{768}
\newcommand{\NumPairingAuditPass}{4,096}

\newcounter{appendixref}

\lstdefinestyle{webriderprompt}{
  basicstyle=\scriptsize\ttfamily,
  numbers=none,
  frame=single,
  framesep=3pt,
  rulecolor=\color{black!25},
  backgroundcolor=\color{black!2},
  xleftmargin=0pt,
  breaklines=true,
  aboveskip=2pt,
  belowskip=2pt,
  showstringspaces=false,
  columns=fullflexible
}

\newenvironment{appxtable}{%
  \begin{table}[!htbp]\centering
}{%
  \end{table}
}

\newenvironment{appxtablewide}{%
  \begin{table*}[!tbp]\centering
}{%
  \end{table*}
}

\newenvironment{appxfigure}{%
  \begin{figure}[!htbp]\centering
}{%
  \end{figure}
}

\newcommand{\AppendixLayout}{%
  \raggedbottom
  \titleformat{name=\section,numberless}
    {\large\bfseries\headingfont\raggedright}{}{0em}{##1}%
  \setcounter{topnumber}{5}%
  \setcounter{bottomnumber}{3}%
  \setcounter{totalnumber}{8}%
  \setcounter{dbltopnumber}{4}%
  \renewcommand{\topfraction}{0.92}%
  \renewcommand{\bottomfraction}{0.85}%
  \renewcommand{\textfraction}{0.07}%
  \renewcommand{\floatpagefraction}{0.72}%
  \renewcommand{\dbltopfraction}{0.92}%
  \renewcommand{\dblfloatpagefraction}{0.72}%
  \setlength{\textfloatsep}{12pt plus 2pt minus 2pt}%
  \setlength{\dbltextfloatsep}{12pt plus 2pt minus 2pt}%
  \setlength{\floatsep}{10pt plus 2pt minus 2pt}%
  \setlength{\dblfloatsep}{10pt plus 2pt minus 2pt}%
  \setlength{\intextsep}{10pt plus 2pt minus 2pt}%
  \captionsetup{font=small,skip=5pt}%
}

\makeatletter
\g@addto@macro\AppendixLayout{%
  \setlength{\@fptop}{0pt}%
  \setlength{\@dblfptop}{0pt}%
  \setlength{\@fpsep}{14pt plus 2pt minus 2pt}%
  \setlength{\@dblfpsep}{14pt plus 2pt minus 2pt}%
}
\makeatother

\title{WebRider: Persona-Conditioned Intent Controllers for Live-Web Assistance}

\correspondingauthor{niatzt@google.com}

\author[1,2,*]{Zhi Li}
\author[1]{Tao Zhou}
\author[1]{Yeqing Li}
\author[1]{Eugene Ie}
\author[2]{Demetri Terzopoulos}

\affil[1]{\thepa{}{}}
\affil[2]{University of California, Los Angeles}
\affil[*]{Work done while interning at Google.}

\begin{abstract}
Delegating a web task involves more than asking a question; it requires transferring a policy: what to verify, how to handle uncertainty, which preferences matter, and when to stop. Yet, current live-web agents are evaluated solely on the final answer, ignoring the policy constraints that define the delegation. A plausible final answer can conceal violations of that policy. Our full live audit reveals this critical gap: a strong controller completes 99.2\% of tasks but honors all policy constraints in only 38.8\% of cases. Finishing does not imply fidelity. WebRider bridges this gap by formalizing the delegated policy as an \emph{intent contract}---an operational record of goals, constraints, evidence obligations, answer form, and task-local persona controls that must hold even as web pages change. WebRider employs a hierarchical architecture: a top-layer controller maintains the contract, a middle layer realizes intentions as guarded executable actions, and a tool layer executes these actions via browser, search, and maps tools. Our benchmark, RiderBench, evaluates this design on 4,096 live-web contracts across 42 public websites, auditing both the internal contract state and the visible user experience to determine if a rollout preserved its policy and if the steps were persona-consistent. The guarded middle interface also serves as a high-quality training signal; an 8B action-policy model trained through this interface outperforms executable-only baselines under a fixed controller. By making the browsing path a first-class object, WebRider enables a system that is auditable, human-judgeable, and learnable without conflating action realization with final-answer decisions. \emph{Dataset URL: \url{hf.co/datasets/WebRider/WebRider}}
\end{abstract}

\begin{document}

\maketitle

\section{Introduction}

A browser agent can appear successful for the wrong reason. It may reach a product page and offer a fluent recommendation after disregarding the return policy the user cared about, or answer a local query after drifting to the wrong city. In these cases, the user delegated not just an endpoint, but a specific way of doing the task.

Consider: \emph{``Find a good refurbished 4K mirrorless camera under \$900.''} The hard constraints are explicit. The task-local policy is not: a trust-first user expects seller reputation, condition, warranty, and returns to be checked; a deal seeker may broaden the search to open-box offers; an uncertainty-averse user may insist on asking before proceeding if compatibility is unknown. Crucially, these are not demographic identities or private memories. They are \emph{control policies} governing evidence gathering, uncertainty handling, ranking, interaction, and stopping. Consequently, two controllers can obey the same request yet require entirely different valid routes.

Preserving such a policy is difficult on the live web. Evidence appears gradually, pages change, access can fail, and reasoning, grounding, clicking, clarification, and stopping are entangled in a single loop. Worse, a final answer can remain plausible even if the controller has deviated from the delegated route. The scientific question is: \emph{How can a browser agent preserve a task-local policy through every visible step while the page state changes?}

\begin{figure}[!t]
\centering
\includegraphics[width=\linewidth]{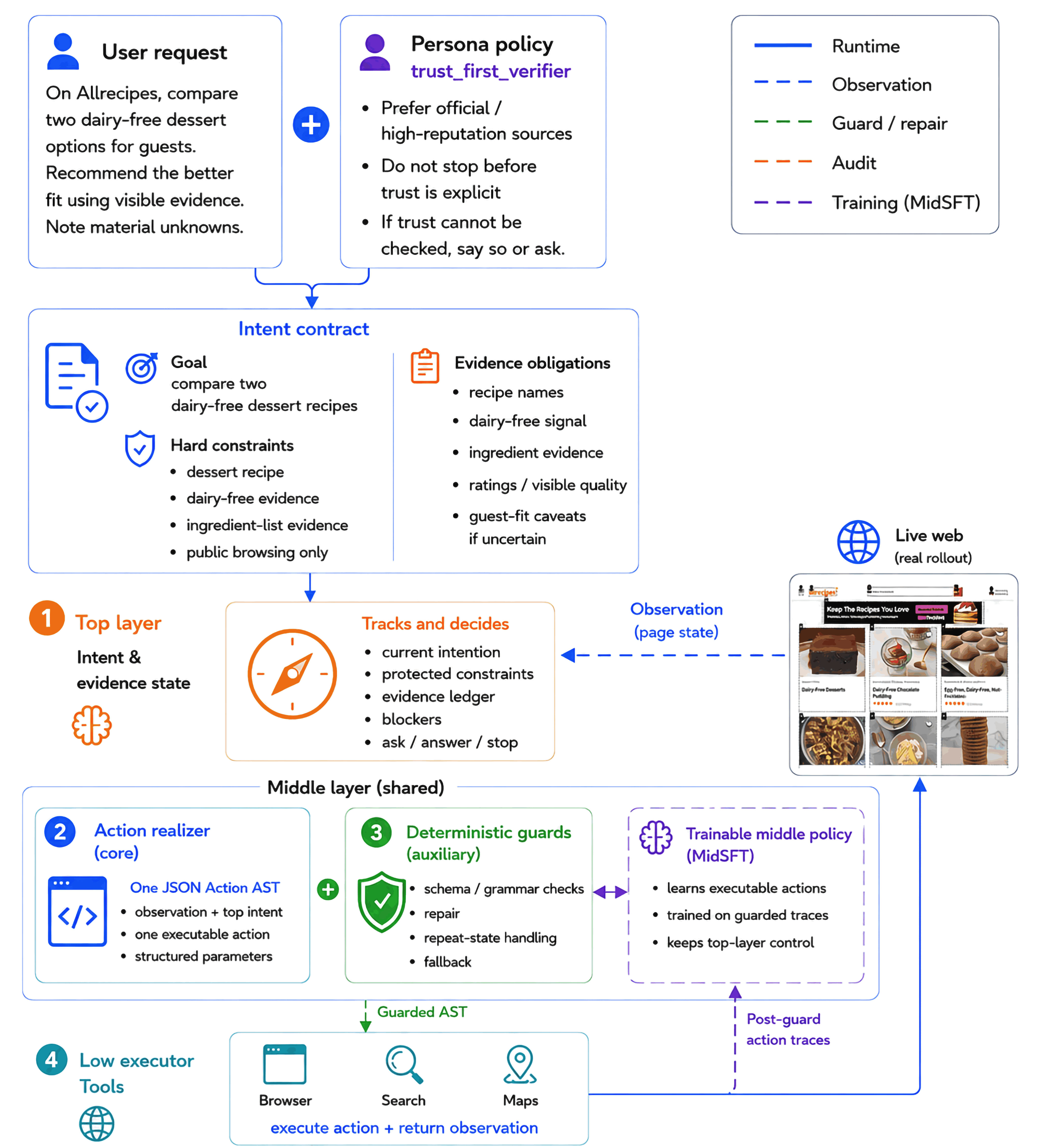}
\caption{The WebRider hierarchy preserves task-local policy. The top layer enforces the intent contract and tracks finalization state; the middle layer generates atomic, guarded executable actions; and the executor layer operates on the live web page, returning observations to the controller.}
\label{fig:architecture}
\end{figure}

We call the stable audit record for that policy an \emph{intent contract}. It encapsulates the goal, constraints, preferences, evidence obligations, blockers, answer form, and persona/control policy. WebRider enforces this record through the hierarchy illustrated in \autoref{fig:architecture}: a top layer tracks the contract and decides whether to browse, ask, stop, block, or answer; a middle layer realizes the current intention as a single, atomic guarded JSON Action AST; and a bottom layer executes browser, search, or maps actions. Persona specifies the policy; the hierarchy preserves it during execution. Significantly, the split is structural, not a claim of cognitive equivalence. It decouples long-horizon sufficiency decisions from page-local action selection and creates a local target whose correctness can be verified after every step.

The paper follows one causal chain. RiderBench expands 768 audited requests into 4,096 persona-conditioned contracts across 42 sites, enabling same-request counterfactuals under fixed hard constraints and evidence obligations. Success is measured via contract gates and a paired human study to evaluate  whether completed traces and visible steps preserve the delegated policy. Finally, ablations isolate whether persistent intention drives control better  than persona or intention text alone, and analysis of guarded middle traces tests whether the resulting action boundary can train a replaceable policy under a fixed controller.

The headline result illustrates why the path matters: Our strong benchmark controller (Full-Pro), terminates on 99.2\% of contracts, but only 38.8\% satisfy every contract gate. The human study finds that final-trace success and stepwise acceptability diverge significantly, while the mechanism study isolates the value of persistent intention. WebRider thus makes missing delegation fidelity observable and provides a controlled way to test, audit, and learn the route that produced the answer.

\section{Related Work}

\paragraph{Web and GUI agents.}
Foundational works such as World of Bits \citep{shi2017worldofbits} and WebShop \citep{yao2022webshop} formulate web interaction as grounded decision making, while Mind2Web \citep{deng2023mind2web} provides broad offline action supervision. More recent efforts, including WebArena \citep{zhou2023webarena} and VisualWebArena \citep{koh2024visualwebarena}, offer controlled executable environments. The field has since expanded to live sites and general computer use with benchmarks like WebVoyager \citep{he2024webvoyager}, BrowserGym \citep{chezelles2024browsergym}, WorkArena \citep{drouin2024workarena}, OSWorld \citep{xie2024osworld}, and BrowserArena \citep{anupam2025browserarena}. Critically, these benchmarks primarily ask whether an agent completes a task or reaches a correct state. By contrast, WebRider adds a task-local delegation record, asking not just \emph{if} the task is done, but whether the policy survives the entire trajectory.

\paragraph{Personalization and persona.}
Recent work on user-conditioned agents (e.g., PersonalWAB \citep{cai2025personalwab}, RealWebAssist \citep{ye2025realwebassist}, Persona2Web \citep{kim2026persona2web}, ShopperBench \citep{ling2026shopperbench}, Orion \citep{hu2026orion}, TIPO \citep{lin2026tipo}, PersonaFingerprint \citep{song2026personafingerprint}, SimPersona \citep{foumani2026simpersona}) aims to move beyond a single average policy. However, existing approaches often conflate persona with stylistic traits or static memory. By contrast, WebRider redefines persona as a controlled counterfactual---an observable, task-local rule governing search breadth, verification, clarification, ranking, and stopping. By pairing these rules with identical requests, site families, hard constraints, and evidence obligations, WebRider isolates useful behavior change from constraint drift or unsupported recommendations.

\paragraph{Modular and trainable control.}
ReAct \citep{yao2023react} interleaves reasoning and action, while SeeAct \citep{zheng2024seeact}, WebGUM \citep{furuta2023webgum}, WebLINX \citep{lu2024weblinx}, Avenir-Web \citep{li2026avenirweb}, and WAC \citep{shen2026wac} improve grounding through visual context, dialogue, memory, experts, or correction. Planning work has likewise separated high-level planning from low-level execution \citep{erol1994htn,sodhi2024heap,aghzal2026hierarchical}. By contrast, WebRider trains a narrower interface: given top intention, persona state, current observation, and recent history, its middle layer predicts a single, post-guard executable Action AST. It does not learn evidence sufficiency or final-answer readiness; these remain top-layer decisions.

\paragraph{Trajectory evaluation and human acceptability.}
Trustworthiness research shows that completion can hide unsupported or unsafe behavior \citep{levy2024stwebagentbench}. Recent benchmarks such as Mind2Web 2 \citep{gou2025mind2web2}, AgentRewardBench \citep{lu2025agentrewardbench}, and Emergence WebVoyager \citep{akkil2026emergencewebvoyager} investigate trajectory judging, ambiguity, and live-web reporting. WebRider goes further by decomposing completion into contract gates, reports access and runtime failures separately, and adds human labels for stepwise persona consistency and delegation comfort. Crucially, model-rater pilots are used only as diagnostics, never as human gold standards.

\section{Hierarchical Controller}
\label{sec:method}

WebRider decouples the decision of \emph{what} still matters from \emph{how} to act on the current web page. Rather than relying on a fixed sequence of LLM calls, the architecture enforces a strict boundary: the top layer manages evidence sufficiency and task finalization, while the middle layer generates a single executable action at a time. This separation ensures that persona and evidence decisions remain inspectable while simultaneously making the action realization process trainable.

\paragraph{Top layer.}
The top layer's state encapsulates the goal, constraints, preferences, blockers, evidence ledger, progress, and next intention. The layer orchestrates the high-level strategy, deciding whether to browse, ask, answer, stop, or flag an external blocker. to continue browsing, the layer emits a compact command, such as ``Verify this candidate's return policy'' or ``Find a second source for compatibility.''

\paragraph{Middle layer and guards.}
Based on the current observation, top-layer command, persona state, and recent history, the middle layer emits exactly one JSON Action AST: \textsc{Click}, \textsc{Type}, \textsc{Scroll}, \textsc{Search}, \textsc{Maps}, \textsc{Back}, \textsc{Wait}, or a safe fallback. This layer grounds the current intention, but it is prohibited from deciding evidence sufficiency or final-answer readiness. A deterministic compiler and a suite of helper guards enforce the grammar, rejecting malformed targets and terminal labels, detecting repeated or blocked states, and attaching recovery actions.
For training, the target is the final post-guard executable action, not raw model text. The same guards remain active when a learned middle layer is deployed, so every browser step supplies a local and verifiable target. Accordingly, the learned policy is strictly scoped to action realization: complexities such as CAPTCHA and access-policy handling, evidence sufficiency, and final answering remain the responsibility of the guards or the top layer.

\begin{figure}[t]
\centering
\includegraphics[width=\linewidth]{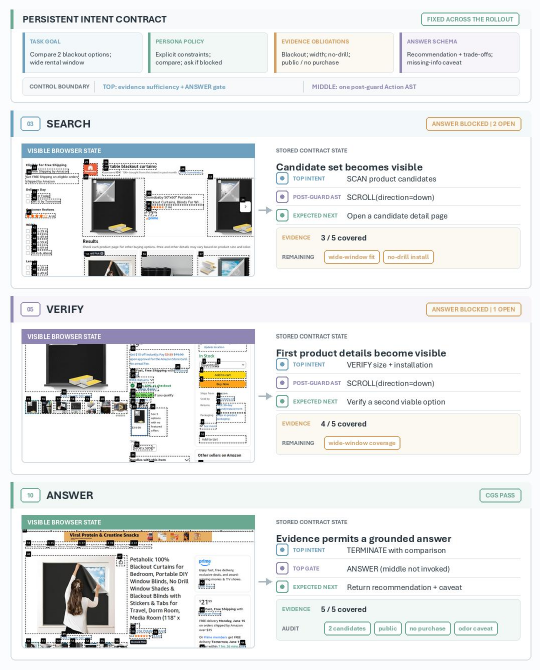}
\caption{An audited rollout pairs each browser state with its stored contract state: current intention, post-guard executable action, expected next observation, evidence coverage, remaining obligations, and final gate. The controller withholds the answer until evidence for blackout performance, width, no-drill installation, and public browsing is covered.}
\label{fig:rollout-audit}
\end{figure}

\paragraph{Executor and trace.}
The executor performs browser and grounding actions, returning screenshots, observations, element labels, and block or readiness signals. It has no persona or evidence reasoning abilities. Each step records the current intention, protected constraints, evidence state, guarded action, and resulting observation. \autoref{fig:rollout-audit} illustrates how these fields make a live trajectory auditable, allowing middle-policy training with the top state, guards, and executor held fixed.

\section{Persona-Conditioned Intent Contracts}
\label{sec:intent-contracts}

An \emph{intent contract} is this paper's operational object for a delegated browsing task. Given request $x$, a website family $w$, and a task-local persona/control policy $p$, the stable record is
\begin{equation}
    (x,w,p) \mapsto (g,H,S,B_{\mathrm{mat}},E,A,p) = z,
\end{equation}
where $g$ denotes the goal, $H$ the hard constraints, $S$ the soft preferences, $B_{\mathrm{mat}}$ the blocking conditions that require asking or stopping, $E$ the evidence obligations, and $A$ the answer schema. Record $z$ changes only upon explicit user clarification, ensuring its obligations are known before interaction and auditable afterward.

The policy specifies how to perform the task, not who the user is. A trust-first verifier checks source credibility (e.g., seller reputation, review count) and returns before recommending; an uncertainty-averse controller asks when an unresolved attribute can change the answer; a deal hunter may broaden the search to include sale, refurbished, or open-box options. Because $H$ and $E$ stay fixed, a policy may alter search breadth, verification depth, ranking, or stopping---but it cannot override a constraint violation or justify an unsupported claim.

At step $t$, the controller top layer maintains state
\begin{equation}
    s^\text{top}_t=f^\text{top}(z,o_{\le t},a_{<t}),
\end{equation}
where $o_{\le t}$ and $a_{<t}$ are the observation and action histories, respectively.
The middle layer maps this state and the current observation $o_t$ to one guarded action
\begin{equation}
    a_t=f^\text{mid}(s^\text{top}_t,o_t,h_t),
\end{equation}
where $h_t$ is the recent action/evidence history. Here $a_t$ is the guarded Action AST executed to produce $o_{t+1}$. Ask, Answer, Blocked, and Stop decisions remain with the top layer; the middle grounds the current intention and may recover through a browser-safe fallback.

This boundary separates \emph{why} the controller acts from \emph{how} it executes its actions. Allowing the middle layer to terminate would entangle click choice with the assessment of evidence sufficiency. Conversely, specifying every click at the top would preclude a reusable action policy. WebRider resolves this by keeping the layers distinct but aligned; the top state records the contractual obligations while the middle layer determines how to advance that intention on the current web page. This separation ensures that the top layer maintains the contract, while the middle layer is the agent of execution.

\paragraph{Contract-gated success.}
We define \emph{Contract-Gated Success} as the logical conjunction of five binary gates:
\begin{equation}
    \cgs = c_T \land c_H \land c_E \land c_A \land \neg c_B,
\end{equation}
where the gates indicate terminal completion $c_T$, hard-constraint satisfaction $c_H$, evidence sufficiency $c_E$, usable answer quality $c_A$, and unresolved material blockers $c_B$. \emph{Aggregate CGS} is the fraction of live rollouts satisfying CGS. Importantly, external access blocks and runtime failures (e.g., CAPTCHAs, browser crashes) are reported separately to ensure they are not conflated with unjustified answering, missed evidence, or constraint drift.

\paragraph{Step-level human acceptability.}
While CGS validates the completed trace, the human audit evaluates whether each visible step preserves the persona policy and whether a rater would trust an agent to execute such behavior. We define these metrics as the \emph{Persona-Consistency Score} (PCS) and \emph{Human-Comfort Preference} (HCP). They complement rather than duplicate CGS; a trace may satisfy final gates while taking intermediate steps that a human delegate would reject.

\begin{figure}[!t]
\centering
\resizebox{0.88\linewidth}{!}{%
\begin{tikzpicture}[font=\scriptsize, node distance=0.30cm, box/.style={draw, rounded corners, thick, align=center, minimum width=1.45cm, minimum height=0.58cm}, arr/.style={-{Latex[length=1.4mm]}, thick}]
\node[box, fill=blue!6] (templates) {Base Tasks\\\NumBaseTasks};
\node[box, fill=purple!6, right=of templates] (persona) {Policies\\\NumPersonaPolicies};
\node[box, fill=green!7, right=of persona] (contracts) {Contracts\\\NumContracts};
\node[box, fill=orange!9, right=of contracts] (rollouts) {Rollout\\Records};
\draw[arr] (templates) -- (persona);
\draw[arr] (persona) -- (contracts);
\draw[arr] (contracts) -- (rollouts);
\end{tikzpicture}}
\caption{RiderBench. Audited base tasks are paired with task-local policies, then executed as live rollout records.}
\label{fig:construction}
\end{figure}

\begin{table}[t]
\centering
\resizebox{\linewidth}{!}{
\begin{tabular}{lr}
\toprule
Measure & Count \\
\midrule
Base tasks & \NumBaseTasks \\
Contracts & \NumContracts \\
Websites / domains & \NumSites{} / \NumDomains \\
Persona policies & \NumPersonaPolicies \\
Train / dev / test & \NumTrainContracts{} / \NumDevContracts{} / \NumTestContracts \\
Base / pairing audit pass & \NumTaskAuditPass/\NumBaseTasks{} / \NumPairingAuditPass/\NumContracts \\
Difficulty 1/2/3/4 & 1024 / 1024 / 1024 / 1024 \\
Ambiguity high/medium/low & 1366 / 1365 / 1365 \\
Task modes & 911 / 909 / 664 / 651 / 651 / 310 \\
Output schemas & 2471 / 1625 \\
Base-task variants 5 / 6 & 512 / 256 \\
\bottomrule
\end{tabular}
}
\caption{RiderBench coverage. Task modes are plan / compare / verify / lookup / select / troubleshoot; outputs are recommendation / evidence-answer.}
\label{tab:dataset-overview}
\end{table}

\begin{figure}[!t]
\centering
\resizebox{\linewidth}{!}{%
\begin{tikzpicture}[x=0.64cm,y=0.45cm,font=\scriptsize,
  dot/.style={circle,draw=black!75,fill=black,inner sep=1.25pt},
  lbl/.style={font=\scriptsize,inner sep=1.0pt},
  quad/.style={font=\scriptsize,align=center,text=black!70}]
\fill[orange!8] (0,0) rectangle (5,5);
\fill[red!6] (5,0) rectangle (10,5);
\fill[blue!6] (0,5) rectangle (5,10);
\fill[green!7] (5,5) rectangle (10,10);
\draw[gray!35] (0,0) rectangle (10,10);
\draw[dashed,gray!70] (5,0) -- (5,10);
\draw[dashed,gray!70] (0,5) -- (10,5);
\draw[-{Latex[length=1.5mm]},semithick] (0,0) -- (10.55,0) node[anchor=west,align=left] {Broader search\\and diversity};
\draw[-{Latex[length=1.5mm]},semithick] (0,0) -- (0,10.55) node[anchor=south,align=center] {Stricter assurance\\and deliberation};
\node[quad] at (2.5,9.35) {cautious direct\\ask/verify};
\node[quad] at (7.5,9.35) {rigorous exploration\\compare evidence};
\node[quad] at (2.5,0.65) {fast/simple\\sufficient stop};
\node[quad] at (7.5,0.65) {opportunity search\\value/deals};

\node[dot] at (1.33,4.20) {}; \node[lbl,anchor=west] at (1.45,4.20) {Urgent};
\node[dot] at (2.67,5.90) {}; \node[lbl,anchor=east] at (2.42,6.18) {Conven.};
\node[dot] at (3.00,5.70) {}; \node[lbl,anchor=east] at (2.83,5.40) {Access};
\node[dot] at (3.33,5.90) {}; \node[lbl,anchor=west] at (3.60,5.70) {Local};
\node[dot] at (3.50,6.00) {}; \node[lbl,anchor=west] at (3.77,6.18) {Brand};
\node[dot] at (4.50,8.40) {}; \node[lbl,anchor=east] at (4.36,8.40) {Safety};
\node[dot,fill=white] at (5.00,5.00) {}; \node[lbl,anchor=north west] at (5.12,4.82) {Control};
\node[dot] at (5.67,5.60) {}; \node[lbl,anchor=west] at (5.92,5.42) {Budget};
\node[dot] at (6.33,4.30) {}; \node[lbl,anchor=west] at (6.45,4.30) {Deal};
\node[dot] at (8.50,5.60) {}; \node[lbl,anchor=west] at (8.62,5.60) {Diversity};
\node[dot] at (6.17,6.20) {}; \node[lbl,anchor=west] at (6.30,6.18) {Sustain.};
\node[dot] at (7.67,6.80) {}; \node[lbl,anchor=west] at (7.80,6.80) {Quality};
\node[dot] at (6.33,7.50) {}; \node[lbl,anchor=west] at (6.45,7.50) {Spec};
\node[dot] at (5.67,8.20) {}; \node[lbl,anchor=west] at (5.80,8.16) {Trust};
\node[dot] at (5.67,8.90) {}; \node[lbl,anchor=east] at (5.45,9.12) {Uncert.};
\end{tikzpicture}}
\caption{Persona-policy projection. Search breadth increases rightward and assurance upward; live rollouts use the full 10-axis policy vector.}
\label{fig:persona-policy-map}
\end{figure}

\begin{table}[!t]
\centering
\resizebox{\linewidth}{!}{
\begin{tabular}{lrrrrrrrrr}
\toprule
Run & $N$ & Term & CGS & Hard & Evid & Ans & Block & RT & Steps \\
\midrule
Full-Pro & 4,096 & 99.2\% & \textbf{38.8\%} & 70.3\% & 57.4\% & 42.3\% & \textbf{13.5\%} & 0.8\% & \textbf{8.37} \\
Full-Flash & 4,096 & 99.7\% & 34.4\% & 68.2\% & 58.6\% & 62.5\% & 15.7\% & \textbf{0.3\%} & 11.49 \\
Flash-60 & 4,096 & 99.0\% & 37.9\% & \textbf{73.3\%} & \textbf{58.8\%} & \textbf{64.1\%} & 17.0\% & 1.0\% & 20.04 \\
Exec-Pro & 4,096 & 99.7\% & 32.9\% & 42.2\% & 35.7\% & 33.0\% & 14.7\% & 0.3\% & 9.09 \\
Exec-Flash & 4,096 & 99.5\% & 32.8\% & 41.3\% & 35.6\% & 33.6\% & 16.9\% & 0.5\% & 11.58 \\
GPT-5.5-Diag & 1,024 & -- & 44.9\% & 67.2\% & 62.5\% & 46.1\% & 3.1\% & 0.0\% & -- \\
\bottomrule
\end{tabular}
}
\caption{Live contract-gated audits. Full-Pro CGS is 38.8\% (95\% CI 37.4--40.3\%). GPT-5.5-Diag is a blocked-site-excluded diagnostic; resource accounting appears only in the supplement.}
\label{tab:main-results}
\end{table}

\begin{figure}[!tb]
\centering
\includegraphics[width=\linewidth]{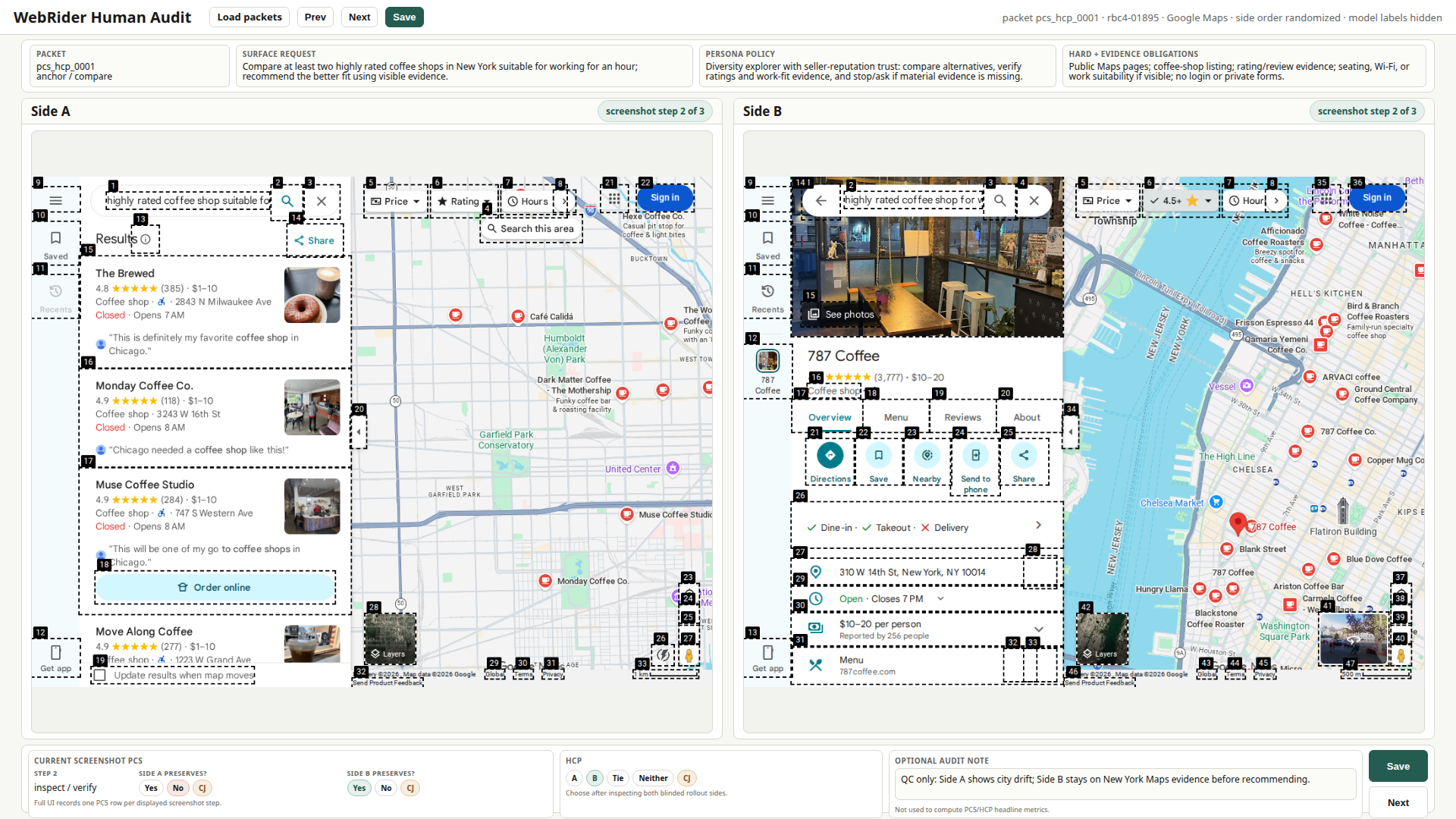}
\caption{Paired human audit. Raters see one contract and two blinded rollouts, then judge persona consistency and delegation comfort. Here, one side drifts to Chicago for a New York request while the other remains on New York evidence.}
\label{fig:human-audit-paired-ui}
\end{figure}

\begin{table*}[!t]
\centering
\resizebox{0.8\linewidth}{!}{
\begin{tabular}{lrrrrrrrr}
\toprule
\multicolumn{9}{l}{\textit{A. Paired human outcome}} \\
Comparison & Pairs & Rollouts & Raters & PCS-A & PCS-B & $\Delta$PCS & HCP net & CJ \\
\midrule
IntentCore vs teacher/full & 80 & 160 & 6 & \textbf{86\%} & 70\% & \textbf{+16 pp} & \textbf{+42.5\%} & $\le$8\% \\
\bottomrule
\end{tabular}
}\\[5pt]
\resizebox{0.8\linewidth}{!}{
\begin{tabular}{lrrrrrrrr}
\toprule
\multicolumn{9}{l}{\textit{B. Agreement and gate calibration}} \\
Diagnostic & Unit $N$ & Raters & $\kappa_H$ & $\kappa_P$ & Exp.-am. $\kappa$ & Exp.-am. acc. & $\rho$/acc. & FP/low \\
\midrule
Anchor PCS/HCP & 20 pairs & 6 & 0.62 & 0.66 & 0.64 & $\ge$85\% & -- & -- \\
PCS--CGS overlap & 160 rollouts & 6 & -- & -- & -- & -- & $\rho\approx0.55$ & 11\% \\
Expert CGS gate & 300 rollouts & 2 & -- & -- & -- & -- & 0.88 acc. & 6.0\% FP \\
\bottomrule
\end{tabular}
}
\caption{Human PCS/HCP outcomes and CGS calibration. Panel~A: CJ combines neither/cannot-judge responses. Panel~B: FP/low reports gate false positives or low-PCS CGS passes.}
\label{tab:human-audit-main}
\end{table*}

\begin{table*}[!t]
\centering
\resizebox{0.8\textwidth}{!}{
\begin{tabular}{llrrrrrr}
\toprule
Condition & Interface & $N$ & CGS & Hard-C & Evid & Ans & Early Ans \\
\midrule
Prompt & no persona/intent fields & 1,024 & 36.5\% & 66.0\% & 52.0\% & 55.0\% & 31\% \\
Prompt+PI & persona+intent in prompt & 1,024 & 41.2\% & 70.5\% & 60.5\% & 62.0\% & 24\% \\
Teacher/full & original controller & 1,024 & 40.0\% & 72.6\% & 56.8\% & 59.3\% & -- \\
Flat & one AST + guards & 1,024 & 44.6\% & 82.4\% & 70.9\% & 71.1\% & $\sim$14\% \\
Persona-only & one AST + persona & 1,024 & 42.6\% & \textbf{84.1\%} & 70.6\% & 70.6\% & -- \\
Intent-only & one AST + intention & 1,024 & 45.5\% & 81.0\% & 69.1\% & \textbf{71.8\%} & -- \\
IntentCore & persona + intention + guards & 1,024 & \textbf{46.8\%} & 78.6\% & \textbf{72.1\%} & 71.6\% & \textbf{$\sim$12\%} \\
IntentCore-30 & same interface, 30-step cap & 1,024 & 42.0\% & 76.5\% & 65.6\% & 68.4\% & -- \\
\bottomrule
\end{tabular}
}
\caption{Mechanism ablations. All rows use 20 steps except the horizon diagnostic IntentCore-30. Persistent intention adds +5.6 pp CGS over Prompt+PI and raises evidence to 72.1\%.}
\label{tab:ablations}
\end{table*}

\section{RiderBench}
\label{sec:riderbench}

RiderBench makes persona-conditioned delegation auditable by structuring data into two components: a \emph{task layer} storing persona-free base tasks and persona-conditioned contracts, and a \emph{rollout layer} capturing live observations, guarded actions, side labels, and calibrated outcomes. The benchmark contains \NumBaseTasks\ requests across \NumSites\ sites and \NumDomains\ domains, expanded into \NumContracts\ unique contracts under \NumPersonaPolicies\ policies, partitioned into \NumTrainContracts\ train, \NumDevContracts\ dev, and \NumTestContracts\ test splits. \autoref{fig:construction} and \autoref{tab:dataset-overview} summarize the construction process and coverage.

Multiple policies are paired with each base task, while hard constraints and evidence obligations remain fixed. This supports same-request counterfactuals: a route may change because the delegated policy changes, but correctness is unachievable by relaxing the task constraints. \autoref{fig:persona-policy-map} projects the 15-policy library onto two axes for visualization, while live rollouts use the full 10-axis vector.

The released step-row schema documents the evidence chain by co-locating the request and constraints, persona policy, top intention, screenshot (or observation), executable action, and verification labels. Thus, trajectories are inspectable both as raw browser behavior and as a record of intent contract preservation.

The audit precedes rollout scoring and excludes tasks with persona names in surface prompts, non-browsing requests, non-inspectable hard constraints, missing evidence obligations, and policies that cannot alter observable behaviors (search, verification, clarification, ranking, or stopping). To ensure rigor, difficulty is exactly balanced, ambiguity varies by at most one contract, and every base task includes 5 to 6 policy variants. These controls enable block-aware live evaluation, pre-specified slice analysis, and counterfactual tests without post-hoc filtering based on success.

Construct validity rests on observable behavior: a policy is meaningful only if it alters the trajectory as intended while preserving hard constraints and evidence. This separation makes persona failure falsifiable, distinguishing it from mere stylistic judgment.

\begin{table*}[!t]
\centering
\resizebox{0.8\linewidth}{!}{
\begin{tabular}{lllrrrrl}
\toprule
\multicolumn{8}{l}{\textit{A. Direct WebRider and human evidence}} \\
Signal & Metric & Comparison & $N$ & Base & Test & $\Delta$ & Auxiliary \\
\midrule
Intention & CGS & teacher $\rightarrow$ IntentCore & 1,024 & 40.0 & \textbf{46.8} & \textbf{+6.7} & Evid/Ans 72.1/71.6 \\
Human & PCS & teacher $\rightarrow$ IntentCore & 80 pairs & 70 & \textbf{86} & \textbf{+16} & HCP +42.5 \\
Persona & route & same-base counterfactuals & 768 groups & -- & 96--98 & -- & spread 35--36 \\
Middle & CGS & Exec-Pro $\rightarrow$ MidSFT-8B & 1,024 & 31.2 & \textbf{50.8} & \textbf{+19.5} & Evid 78.1 \\
\bottomrule
\end{tabular}
}\\[5pt]
\resizebox{0.8\linewidth}{!}{
\begin{tabular}{lllrrrrl}
\toprule
\multicolumn{8}{l}{\textit{B. External proxy checks}} \\
Signal & Metric & Proxy & $N$ & Base & Test & $\Delta$ & $p$ \\
\midrule
Intention & step succ. & Mind2Web semantic & 1,339 & 30.8 & \textbf{42.7} & \textbf{+11.9} & $<10^{-12}$ \\
Intention & step succ. & WebLINX dialogue & 500 & 34.8 & \textbf{39.0} & \textbf{+4.2} & 0.0275 \\
Interface & step succ. & Mind2Web screenshot+state & 500 & 36.8 & \textbf{43.6} & \textbf{+6.8} & $7.56{\times}10^{-5}$ \\
Persona & gate pass & PersonalWAB-style & 120 & 0.0 & \textbf{15.0} & \textbf{+15.0} & $7.63{\times}10^{-6}$ \\
\bottomrule
\end{tabular}
}
\caption{Direct and external evidence. Panel A reports live, human, counterfactual, and learned-middle results; Panel B reports related proxy tasks.}
\label{tab:three-axis-baselines-main}
\end{table*}

\begin{figure*}[t]
\centering
\includegraphics[width=0.9\linewidth]{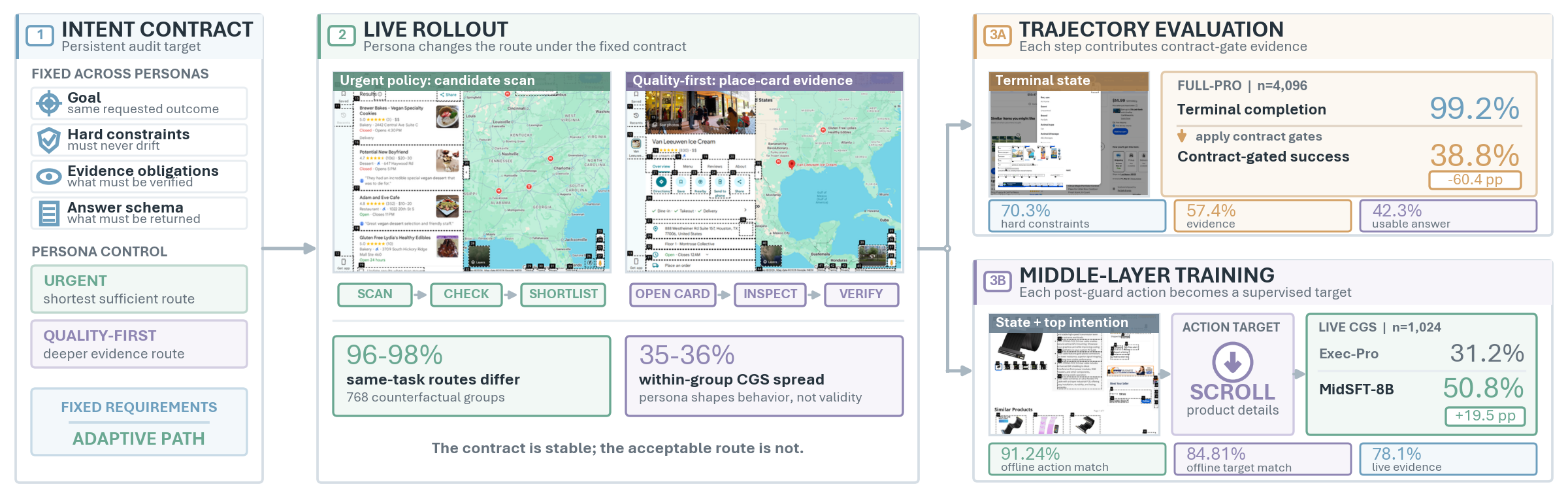}
\caption{One guarded trace supports both trajectory evaluation and middle-layer training. The contract remains the audit target as persona changes the route; the complete trace accumulates contract-gate evidence, while each step yields a supervised example whose target is the post-guard executable Action AST. Callouts summarize route divergence, contract-gate outcomes, and learned-middle performance.}
\label{fig:claim-map}
\end{figure*}

\section{Experiments and Evaluation}
\label{sec:experiments}

We evaluated four key aspects: (1) whether completed live traces preserved their contracts, (2) whether humans accepted the visible steps, (3) whether structure added value beyond persona/intention text, and (4) whether the guarded middle action was learnable. All live runs utilized the same 4,096-contract dataset, a Chromium backend, and a calibrated audit protocol ensuring at most one active task per site. External blocks and runtime failures were labeled separately to distinguish infrastructure issues from policy violations. When page-local evidence was unavailable, blocked, or geographic, Search and Maps tools were audited for grounding accuracy. 

Evidence and answer gates combined deterministic checks with judgments from GPT-5.5 (high-reasoning) and Gemini 3.1 Pro (preview), calibrated against a two-expert audit of 300 rollouts. Paired mechanism evaluations used the same contract slice and a 20-step budget unless specified as a horizon diagnostic. 

We compared four model configurations: Full-Pro/Full-Flash (complete controllers), Exec-Pro/Exec-Flash (restricted to executable actions in the middle layer), IntentCore (binding persona and top intention to a single guarded action), and MidSFT-8B (replacing only the middle policy). The supplement maps these names to archived run identifiers, launch settings, and training recipes.

\paragraph{Live contract-gated evaluation.}
\autoref{tab:main-results} disentangles termination from contract quality, access blocks, and runtime failures. While GPT-5.5-Diag excludes heavily blocked sites and thus does not constitute a matched full-run comparison, the main results reveal a critical gap: 
Full-Pro terminated on 99.2\% of contracts but passed all gates on only 38.8\%. This highlights that evidence and answer gates identified many trajectories that terminated without sufficient evidence.

\paragraph{Stepwise human acceptability.}
Raters compared two blinded rollouts for the same contract, marking each side's persona consistency and choosing the behavior towards which they would rather delegate (\autoref{fig:human-audit-paired-ui}). Arm A was IntentCore and Arm B was the teacher/full controller. Six raters labeled 80 pairs (160 rollouts): 20 shared anchors supported agreement estimates, and 60 coverage pairs were labeled by two amateur raters after expert calibration. A separate two-expert audit calibrated CGS on 300 rollouts.
The outcomes in \autoref{tab:human-audit-main} show that IntentCore achieved 86\% PCS versus 70\% for teacher/full controller, yielding a +42.5\% net HCP advantage. Notably, PCS is correlated with but distinct from CGS ($\rho\approx0.55$).

\paragraph{Hierarchy adds more than information.}
\autoref{tab:ablations} compares prompt-only and structured controllers on the paired 1,024-contract slice. Prompt+PI supplied the same persona and intention information as text, whereas IntentCore bound this information to a persistent top state and a single guarded action. The matched gain was +5.6 percentage points (pp) in CGS, while the improvement from the teacher/full controller to IntentCore was +6.7\,pp (95\% paired CI: +3.9\,pp to +9.5\,pp; 143 vs. 74 favorable transitions).

\paragraph{Convergent evidence.}
\autoref{tab:three-axis-baselines-main} distinguishes direct WebRider evidence from external proxy checks. The proxy baselines test related action-grounding or profile-conditioning signals and are not entries on the live WebRider leaderboard.

\paragraph{Learning the action-realization layer.}
While the preceding ablations tested whether hierarchy improves control, this experiment asked whether its per-step interface is reusable. The top layer handled long-horizon evidence and stopping decisions, and the executor was deterministic; thus, only the middle layer, mapping local states to post-guard actions, required replacement. \autoref{tab:trained-middle-main} reports results for both held-out prediction and live insertion. MidSFT-8B achieved 91.24\% action match and 84.81\% target match, along with 50.8\% CGS and 78.1\% evidence on 1,024 live contracts. This validates the shared evaluation and training boundary shown in \autoref{fig:claim-map}, rather than requiring full-controller replacement.

\begin{table}[t]
\centering
\resizebox{\linewidth}{!}{
\begin{tabular}{lrrrrr}
\toprule
Middle Result & $N$ & Action & Target & CGS & Evid \\
\midrule
MidSFT-8B offline & 2,146 & \textbf{91.24\%} & \textbf{84.81\%} & -- & -- \\
Qwen8B-Exec offline & 2,422 & 88.93\% & 82.33\% & -- & -- \\
Qwen32B offline & 1,024 & 90.23\% & 82.81\% & -- & -- \\
Gemma4-31B offline & 1,024 & 72.27\% & 79.69\% & -- & -- \\
Qwen30B-A3B offline & 128 & 82.81\% & 75.78\% & -- & -- \\
Exec-Pro ref. live & 1,024 & -- & -- & 31.2\% & -- \\
Exec-Flash ref. live & 1,024 & -- & -- & 34.8\% & -- \\
MidSFT-8B live & 1,024 & -- & -- & \textbf{50.8\%} & \textbf{78.1\%} \\
\bottomrule
\end{tabular}
}
\caption{Trained middle layer results. Offline rows predict post-guard actions, while live rows hold the Gemini top layer and guards fixed.}
\label{tab:trained-middle-main}
\end{table}

\paragraph{Results and discussion.}
The first result is a gap between termination and fidelity. Full-Pro almost always stopped, yet less than 40\% of traces satisfied all contract gates (\autoref{tab:main-results}). Hard constraints passed more often than evidence and answer quality, so the missing mass was not merely browser failure: many completed traces reached a plausible endpoint without assembling the support or preserving the conditions required by the contract. Contract-gated scoring exposed this failure while retaining block and runtime labels as separate causes.

Human judgments revealed a complementary gap. IntentCore's 86\% PCS and +42.5\% net HCP advantage aligned with its automatic improvement, yet PCS and CGS correlated only moderately; 11\% of CGS-passing human-audit rollouts were assigned low PCS scores. The two measurements therefore answer different questions: CGS asks whether the completed trace satisfies auditable obligations, while PCS/HCP asks whether the visible behavior remains acceptable to a person delegating under that policy.

The ablations pinpoint the mechanism more narrowly. Persona and intention text improved a prompt-only controller, but adding a persistent top state plus one guarded action yielded further improvement (\autoref{tab:ablations}). The gain was not reproduced by a longer horizon: IntentCore-30 fell to 42.0\% CGS and incurred more runtime failures. Same-base counterfactuals supported the intended role of the persona: 96--98\% of 768 groups changed action signature and roughly 35--36\% showed within-group success spread. Persona is thus a behavior-shaping control variable, not a guarantee of higher success or a change to the task's hard requirements.

The executable-only runs clarified the learning result. Removing finalization from the middle lowered Full-Pro CGS from 38.8\% to 32.9\%, showing that a clean interface is not automatically a better live controller. It did, however, define a local target that MidSFT-8B learned well. Under the fixed top layer, the student reached 50.8\% CGS and 78.1\% evidence, surpassing the same-subset Exec-Pro/Exec-Flash references. Because access pressure differed across runs, \autoref{tab:block-adjusted-main} also reports no-block and clean denominators; the clean CGS was 52.0\% versus 38.9\%/39.9\%. This serves as evidence for the trainability of the action-realization boundary, not a full-agent model-family ranking.

\begin{table}[t]
\centering
\resizebox{0.8\linewidth}{!}{
\begin{tabular}{lrrrr}
\toprule
Run & All & Block & No-block & Clean \\
\midrule
Exec-Pro & 32.9 & 14.7 & 38.8 & 38.9 \\
Exec-Flash & 32.8 & 16.9 & 39.6 & 39.9 \\
MidSFT-8B & 50.8 & 2.0 & 52.0 & 52.0 \\
\bottomrule
\end{tabular}
}
\caption{Block-adjusted CGS (\%). No-block removes access blocks; Clean further excludes runtime failures.}
\label{tab:block-adjusted-main}
\end{table}

\begin{table}[t]
\centering
\resizebox{0.8\linewidth}{!}{
\begin{tabular}{lrrr}
\toprule
Slice & Groups & CGS Span & Block Span \\
\midrule
Website & 42 & 0.0--90.5 & 0.0--100.0 \\
Domain & 12 & 9.4--68.8 & 0.0--34.4 \\
Task type & 6 & 30.4--46.6 & 12.7--14.9 \\
Difficulty & 4 & 37.3--40.3 & 11.9--14.7 \\
\bottomrule
\end{tabular}
}
\caption{Full-Pro cross-slice ranges over 4,096 contracts. Variation across websites and domains exceeds the baseline difficulty spread.}
\label{tab:cross-slice-main}
\end{table}

The slice analysis identified where live-web difficulty enters. Full-Pro spanned 0.0--90.5\% CGS across sites but only 37.3--40.3\% across the balanced author difficulty levels (\autoref{tab:cross-slice-main}). Task form mattered even at similar block rates: Compare tasks reached 46.6\% CGS, whereas Select tasks reached 30.4\%. These results refined the aggregate claim in a useful way. WebRider measures contract preservation under changing public-web conditions, where site affordance, access policy, evidence availability, and stopping structure can dominate a nominal difficulty tag.

External proxies serve only as scope checks: intention helped on Mind2Web and WebLINX, state aided visual grounding, and persona gating altered PersonalWAB-style outcomes.

\section{Conclusions and Future Work}

Persona-conditioned browsing is fundamentally a delegation problem: constraints, evidence thresholds, and rules for asking or stopping determine which route is acceptable. WebRider makes that route explicit through intent contracts, audits it at both trace and step levels, and employs a hierarchy that keeps finalization decisions above executable action selection. RiderBench reveals three critical insights: (1) high termination can coexist with low contract fidelity, (2) human judgments confirm that final gates do not fully capture stepwise acceptability, and (3) guarded middle traces provide a learnable action interface under a fixed top layer. For this setting, the path is part of the task, and explicit intent makes it possible to verify whether that path was preserved.

\paragraph{Limitations and Future Work.}

WebRider studies given task-local policies, not private long-term memory or contract inference from dialogue; policy elicitation remains future work. The hierarchy is structural, not a claim of cognitive equivalence. Live sites change, and CAPTCHA or access-policy blocks remain unresolved. We report access and runtime failures separately, though later reruns may drift. Baseline coverage is strongest for Gemini controllers, and GPT-5.5-Diag was not a matched full run.

Evidence and answer gates combine GPT-5.5 high-reasoning and Gemini 3.1 Pro preview judgments with deterministic checks, calibrated on a two-expert, 300-rollout audit. These are less mechanical than termination or block detection. The 80-pair PCS/HCP study blinds raters to run names, but rollout style can reveal cues; not every mechanism contrast has human labels (see supplement for HCP tie normalization). MidSFT-8B replaces only the action-realization middle layer under a fixed top layer and active guards. Full-controller learning and a full 4,096-contract student run remain future work.

\paragraph{Ethical Considerations.}

RiderBench uses non-logged-in public-web tasks and excludes purchases, credentials, private accounts, irreversible actions, and sensitive flows. Persona policies are abstract task-local rules rather than demographic profiles. The release distinguishes between authored contracts, model-generated text, public-web observations, and human labels. Screenshot-reduced artifacts serve as the fallback when website terms or third-party content require it. Operational traces do not expose hidden model reasoning.

\paragraph{Reproducibility.}

The release centers on the atomic unit of execution: one browser step under a single contract. It includes audited contracts, code, manifests, a 100-task bundle with 827 screenshots, and step splits for Full-Pro, Full-Flash, Flash-60, Exec-Pro, and Exec-Flash. Data rows store the request, constraints, persona, top intention, observation, guarded action, verification, and outcome; rollout identifiers and step indices reconstruct trajectories and metrics. The supplement documents launch, training, human-audit, table-regeneration, and release procedures. Credentials, private state, tokens, raw rater material, and restricted third-party content are excluded.

{\small\bibliography{references}}
\flushcolsend
\clearpage
\raggedcolsend

\FloatBarrier
\AppendixLayout
\section{Appendix A: Guide to the Evidence}
\refstepcounter{appendixref}
\label{app:menu}

The appendix follows the order in which a reader would reconstruct the study. \autoref{app:task-audit} specifies the benchmark tasks and intent contracts; \autoref{app:persona-policies} defines the task-local policies and counterfactual pairing; \autoref{app:experimental-protocol} gives the evaluation and rollout protocol; \autoref{app:trained-middle} documents middle-policy training; \autoref{app:fullscale-diagnostics} expands the quantitative results; \autoref{app:human-audit} reports the human study; \autoref{app:runtime-prompts} records the prompt and executable-action interfaces; \autoref{app:robustness} collects uncertainty, access, stratified, and resource analyses; \autoref{app:examples-failures} provides worked cases; and \autoref{app:release-notes} states the release boundary.

\paragraph{Reading order.}
Readers checking benchmark validity can begin with \autoref{app:task-audit} and \autoref{app:persona-policies}. Reproduction of the main comparisons starts in \autoref{app:experimental-protocol}, with learned-middle details in \autoref{app:trained-middle}. The quantitative and audit evidence follows in \autoref{app:fullscale-diagnostics}--\autoref{app:robustness}. Each section states the setup before its tables and closes with the conclusion supported by them.

\section{Appendix B: Tasks and Intent Contracts}
\refstepcounter{appendixref}
\label{app:task-audit}

RiderBench separates the public task from the policy under which it is delegated. A base task fixes the web objective, hard constraints, evidence obligations, material blockers, and requested answer form. Pairing adds a task-local persona/control policy without changing those auditable requirements. \autoref{tab:contract-schema} shows the representation used throughout the paper.

\begin{appxtable}
\footnotesize
\begin{tabularx}{\linewidth}{@{}>{\raggedright\arraybackslash}p{1.95cm}X@{}}
\toprule
Field & Camera example \\
\midrule
Request & Find a refurbished 4K mirrorless camera under \$900. \\
Goal & Recommend a valid, purchasable candidate. \\
Hard constraints & Refurbished; 4K; price below \$900. \\
Preferences & Seller reputation and return confidence. \\
Blockers & Compatibility or coverage remains unknown. \\
Evidence & Price, condition, feature, seller, and return-policy evidence. \\
Answer form & Selection, support, caveat, and alternatives. \\
Persona policy & Trust-first verification before recommendation. \\
\bottomrule
\end{tabularx}
\caption{Intent-contract schema. Constraints, preferences, blockers, evidence obligations, and answer form remain independently inspectable.}
\label{tab:contract-schema}
\end{appxtable}

The task audit runs before any rollout is scored. It rejects persona names in the surface request, non-browsing tasks, non-inspectable hard constraints, missing evidence requirements, and pairings in which the policy cannot change search, verification, ranking, asking, or stopping. \autoref{tab:task-audit-validation} reports the deterministic checks.

\begin{appxtable}
\footnotesize
\begin{tabular}{@{}lrr@{}}
\toprule
Check & Observed & Required \\
\midrule
Base tasks & 768 & 768 \\
Websites / domains & 42 / 12 & 42 / 12 \\
Contracts & 4,096 & 4,096 \\
Task-audit pass & 768 & 768 \\
Pairing-audit pass & 4,096 & 4,096 \\
Schema issues & 0 & 0 \\
Persona names in requests & 0 & 0 \\
\bottomrule
\end{tabular}
\caption{Final deterministic task and pairing audit, completed before live outcomes were available.}
\label{tab:task-audit-validation}
\end{appxtable}

The pre-specified task strata are summarized in \autoref{tab:dataset-distribution-app}. Difficulty is exactly balanced, ambiguity differs by at most one contract, and every domain contributes 64 base tasks. These controls support later slice analyses without post-hoc filtering.

\begin{appxtable}
\footnotesize
\begin{tabularx}{\linewidth}{@{}p{1.45cm}Xr@{}}
\toprule
Dimension & Category & Contracts \\
\midrule
\multirow{4}{*}{Difficulty} & Level 1 & 1,024 \\
 & Level 2 & 1,024 \\
 & Level 3 & 1,024 \\
 & Level 4 & 1,024 \\
\addlinespace[1.5pt]
\multirow{3}{*}{Ambiguity} & High & 1,366 \\
 & Medium & 1,365 \\
 & Low & 1,365 \\
\addlinespace[1.5pt]
\multirow{6}{*}{Task mode} & Plan & 911 \\
 & Compare & 909 \\
 & Verify & 664 \\
 & Lookup & 651 \\
 & Select & 651 \\
 & Troubleshoot & 310 \\
\addlinespace[1.5pt]
\multirow{2}{*}{Output} & Recommendation & 2,471 \\
 & Evidence answer & 1,625 \\
\addlinespace[1.5pt]
\multirow{3}{*}{Split} & Train & 2,560 \\
 & Development & 512 \\
 & Test & 1,024 \\
\bottomrule
\end{tabularx}
\caption{Contract distribution across pre-specified evaluation strata and the base-task-grouped learning split.}
\label{tab:dataset-distribution-app}
\end{appxtable}

\section{Appendix C: Persona Policies and Counterfactual Pairing}
\refstepcounter{appendixref}
\label{app:persona-policies}

Here, \emph{persona} denotes a task-local decision policy over search breadth, assurance, cost, uncertainty, locality, and stopping while the base request and hard constraints remain fixed. \autoref{tab:persona-axis-app} gives the construction axes.

\begin{appxtable}
\footnotesize
\begin{tabularx}{\linewidth}{@{}p{1.35cm}X@{}}
\toprule
Axis & High setting \\
\midrule
Cost & prioritize total price, fees, and discounts among valid options \\
Trust & prefer official, reputable, and auditable sources \\
Speed & stop once safe evidence is sufficient \\
Evidence & verify each material claim before answering \\
Clarify & ask when an unresolved fact can change the decision \\
Explore & compare candidates and sources before settling \\
Privacy & avoid unnecessary accounts, disclosure, or irreversible actions \\
Access & prefer readable, stable, low-friction interaction paths \\
Locality & use location, hours, dates, availability, and travel time \\
Diversity & surface meaningfully different valid options \\
\bottomrule
\end{tabularx}
\caption{Persona-policy axes. Intermediate values encode balanced settings between the endpoints.}
\label{tab:persona-axis-app}
\end{appxtable}

Near-neighbor policies share several tendencies but must differ on one auditable decision. \autoref{tab:persona-neighbor-contrasts-app} records the discriminants used to validate same-task counterfactuals.

\begin{appxtable}
\footnotesize
\begin{tabularx}{\linewidth}{@{}p{2.1cm}X@{}}
\toprule
Contrast & Required distinction \\
\midrule
Budget / deal & total-value accounting vs. sale, refurbished, or open-box discovery \\
Convenience / access & resolving confusion vs. preferring stable, readable interfaces \\
Trust / spec & source authority vs. exact technical compatibility \\
Trust / uncertainty & accept after strong evidence vs. ask or abstain under material ambiguity \\
Quality / sustainability & quality fit vs. durability, repairability, locality, or waste reduction \\
Privacy / uncertainty & sensitive-flow risk vs. epistemic uncertainty \\
\bottomrule
\end{tabularx}
\caption{Near-neighbor counterfactuals. A pair is valid only when the distinction can change the route without relaxing hard constraints.}
\label{tab:persona-neighbor-contrasts-app}
\end{appxtable}

\paragraph{Counterfactual trajectory statistics.}
For each base task, we group rollouts that differ only in persona policy. An
\emph{action signature} is the ordered sequence of normalized action kinds
(e.g., \textsc{type--scroll--click--answer}); a group is action-divergent when
it contains more than one signature. Its diversity index is
$1-\max_s n_s/n$, where $n_s$ is the frequency of signature $s$. A group has
a CGS spread when it contains both a passing and a non-passing rollout.
\autoref{tab:persona-counterfactual-app} reports these descriptive quantities;
they establish behavioral sensitivity to policy while PCS/HCP evaluates
whether the changed behavior is preferable.

\begin{appxtable}
\footnotesize
\setlength{\tabcolsep}{2.1pt}
\begin{tabular}{@{}lrrrrr@{}}
\toprule
Run & Groups & \makecell[r]{Unique\\sig.} & Divergent & Diversity & \makecell[r]{CGS\\spread} \\
\midrule
Full-Pro & 768 & 4.30 & 96.5\% & 0.635 & 35.4\% \\
Full-Flash & 768 & \textbf{4.65} & \textbf{97.9\%} & \textbf{0.694} & \textbf{36.3\%} \\
\bottomrule
\end{tabular}
\caption{Same-base-task counterfactuals over all 4,096 contracts per run (5.33 policies per group). Among 8,960 within-group policy pairs, 14.6\% (Pro) and 14.9\% (Flash) switch between CGS pass and non-pass.}
\label{tab:persona-counterfactual-app}
\end{appxtable}

\autoref{tab:persona-full} lists all 15 policies. Each row includes a behavioral criterion that can fail on an observed trajectory, making the policies observable controls over action choice.

\begin{appxtable}
\footnotesize
\begin{tabularx}{\linewidth}{@{}>{\raggedright\arraybackslash}p{1.95cm}r>{\raggedright\arraybackslash}X@{}}
\toprule
Policy & $N$ & Behavioral criterion \\
\midrule
Explicit control & 768 & follow stated constraints; no latent preference may redirect the route \\
Trust-first verifier & 446 & prefer authoritative evidence; extra browsing must strengthen support \\
Spec-exact researcher & 375 & verify exact compatibility and technical attributes \\
Quality-first explorer & 298 & compare observable quality signals before ranking \\
Local-context planner & 275 & use distance, hours, dates, and availability \\
Urgent pragmatist & 274 & stop after sufficient safe evidence without skipping blockers \\
Uncertainty-averse & 271 & ask or abstain when ambiguity can change the result \\
Brand loyalist & 248 & honor an explicit brand preference without inventing one \\
Safety/privacy guardian & 224 & avoid accounts, credentials, purchases, and private data without permission \\
Diversity explorer & 200 & return distinct valid alternatives without relaxing constraints \\
Deal hunter & 164 & search sale/refurbished/open-box options and verify total cost \\
Sustainability-minded & 149 & seek durability, repairability, refurbished, or local evidence \\
Accessibility seeker & 138 & prefer robust, readable, low-cognitive-load paths without reducing correctness \\
Convenience novice & 137 & prefer clear, low-friction interaction and clarify confusing steps \\
Budget optimizer & 129 & rank valid options by total value, including fees and add-ons \\
\bottomrule
\end{tabularx}
\caption{Complete task-local persona-policy library. Counts are contract counts; every row states an observable criterion for the route.}
\label{tab:persona-full}
\end{appxtable}

\section{Appendix D: Evaluation and Experimental Protocol}
\refstepcounter{appendixref}
\label{app:experimental-protocol}

This section fixes the scoring rules, controller boundaries, launch settings, and comparison discipline used by every quantitative result. \autoref{tab:metrics-app} separates contract failures from access and infrastructure failures.
Throughout the paper, \emph{external-access failure} and the compact
\emph{Block}/\emph{No-block} columns refer to the environmental variable
$X_{\mathrm{access}}$; \emph{task blocker} refers only to the material
feasibility condition $B_{\mathrm{mat}}$ in the intent contract. The two terms
are never used interchangeably.

\begin{appxtable}
\footnotesize
\begin{tabularx}{\linewidth}{@{}p{1.6cm}X@{}}
\toprule
Measure & Definition \\
\midrule
CGS & terminal rollout satisfying hard constraints, evidence, answer quality, and material-blocker checks \\
Hard-C & all mandatory constraints are satisfied \\
Evidence & required page or source evidence is observed before finalization \\
Answer & final response is usable and follows the requested form \\
Access $X_{\mathrm{access}}$ & CAPTCHA, login wall, denial, or anti-bot interruption; reported outside CGS components \\
Runtime & crash, timeout, or infrastructure failure; reported separately \\
Persona consistency & human judgment that visible steps preserve the task-local policy \\
Human comfort & paired preference for the rollout that is safer or more comfortable to delegate \\
\bottomrule
\end{tabularx}
\caption{Evaluation measures. Automatic gates score the completed trace; human measures evaluate the visible route under the same contract.}
\label{tab:metrics-app}
\end{appxtable}

\paragraph{Persona coverage.}
CGS has no separate positive-persona conjunct. Persona-derived requirements that
become task obligations enter Hard-C or Evidence, while the automatic persona
proxy detects explicit policy contradictions. Positive route-level fidelity---
including search breadth, verification depth, clarification, and stopping---is
measured by the stepwise PCS/HCP study in \autoref{app:human-audit}.

The hierarchy also determines what a fair mechanism comparison holds fixed. The top layer owns long-horizon evidence and finalization; the executor is deterministic; the middle produces one local, post-guard action target at each browser step. \autoref{tab:controller-boundary-app} summarizes the resulting boundaries.

\begin{appxtable}
\footnotesize
\begin{tabularx}{\linewidth}{@{}p{1.45cm}X@{}}
\toprule
Family & Boundary and held-fixed components \\
\midrule
Prompt controls & prompt-local reasoning/finalization; browser and evaluation stack fixed \\
Full & persistent top state; API middle may propose browser or terminal actions; guards/executor fixed \\
Exec & top owns terminal decisions; middle emits one executable Action AST \\
IntentCore & persistent persona-conditioned intention feeds one guarded action; contract slice, model, and budget fixed \\
MidSFT & Gemini top, guards, executor, and grammar fixed; only the action policy is learned \\
\bottomrule
\end{tabularx}
\caption{Controller boundaries used by the main comparisons.}
\label{tab:controller-boundary-app}
\end{appxtable}

The descriptive variant names encode the boundary under test. \autoref{tab:run-names-app} gives the backbone, evaluated scope, and step cap. Prompt, structural, and executable-boundary comparisons use the same 20-step cap; Flash-60 and IntentCore-30 are horizon diagnostics rather than architecture comparisons.

\begin{appxtable}
\footnotesize
\setlength{\tabcolsep}{2.6pt}
\begin{tabular}{@{}lcrr@{}}
\toprule
Variant & Backbone & $N$ & Cap \\
\midrule
Full-Pro & Gemini 3.1 Pro & 4096 & 20 \\
Full-Flash & Gemini 3.5 Flash & 4096 & 20 \\
Flash-60 & Gemini 3.5 Flash & 4096 & 60 \\
Exec-Pro & Gemini 3.1 Pro & 4096 & 20 \\
Exec-Flash & Gemini 3.5 Flash & 4096 & 20 \\
Prompt / Prompt+PI & Gemini 3.1 Pro & 1024 & 20 \\
Structural arms & Gemini 3.1 Pro & 1024 & 20 \\
IntentCore-30 & Gemini 3.1 Pro & 1024 & 30 \\
GPT-5.5-Diag & GPT-5.5 & 1024 & 20 \\
MidSFT-8B live & Qwen3-VL-8B middle & 1024 & 20 \\
\bottomrule
\end{tabular}
\caption{Experiment variants. GPT-5.5-Diag excludes five heavily access-blocked sites; MidSFT-8B keeps the Gemini top layer, guards, grammar, and executor fixed.}
\label{tab:run-names-app}
\end{appxtable}

Provider throughput differed with available serving capacity, while the scheduler allowed at most one active rollout per website. Worker/RPM values describe collection throughput under that scheduler. \autoref{tab:experiment-setup-app} records the launch settings needed to reproduce each run family.

\begin{appxtable}
\footnotesize
\setlength{\tabcolsep}{2.4pt}
\begin{tabular}{@{}lccc@{}}
\toprule
Run & Workers/RPM & Timeout & Seed/date \\
\midrule
Full-Pro & 8/25 & 900s & 20260518 \\
Full-Flash & 8/60 & 900s & 20260518 \\
Flash-60 & 12/120 & 1800s & 20260525 \\
Structural arms & 8/25 & 900s & 20260521 \\
IntentCore-30 & 8/25 & 1200s & 20260525 \\
Exec-Pro & 6/25 & 900s & 20260606 \\
Exec-Flash & 16/500 & 900s & 20260606 \\
GPT-5.5-Diag & 4/-- & 900s & 20260612 \\
MidSFT-8B live & --/-- & 900s & 20260612 \\
\bottomrule
\end{tabular}
\caption{Live collection settings. Throughput parameters reflect provider capacity under one-active-rollout-per-site scheduling.}
\label{tab:experiment-setup-app}
\end{appxtable}

\paragraph{Controller development settings.}
We did not run a Cartesian hyperparameter sweep.  Architecture choices were
tested as named ablations, while generation and runtime settings were changed
one factor at a time in smoke runs.  Selection never used the final test CGS:
quality-facing settings were fixed before the reported run, and throughput
settings were selected from provider limits and runtime stability.  The exact
development values and decision rules are in
\autoref{tab:controller-development-app}; training choices are reported
separately in \autoref{tab:training-development-app}.  Protocol fields not
listed as varied in the development table remained at the single values in
\autoref{tab:experiment-setup-app} and were not tuned.

\begin{appxtablewide}
\small
\begin{tabularx}{\linewidth}{@{}>{\raggedright\arraybackslash}p{2.45cm}>{\raggedright\arraybackslash}p{4.15cm}X@{}}
\toprule
Parameter & Values tried (count) & Reported choice and selection criterion \\
\midrule
Step cap & 20, 30, 60 (3) & 20 for matched controller comparisons; 30 and 60 retained only as horizon diagnostics \\
Task timeout & 900, 1200, 1800\,s (3) & 900\,s at 20 steps, scaled to 1200/1800\,s for the longer diagnostic horizons \\
Gemini thinking & provider \texttt{high}; explicit 2048-token fallback (2 forms) & \texttt{high} throughout; 2048 makes the same setting explicit when an SDK accepts a budget instead of a level \\
JSON output ceiling & component defaults 500, 600, 1200, 1800, 2400; production 8192 (6) & 8192 for Exec runs after smoke tests exposed truncated Action AST replies; component defaults for earlier Full runs \\
Sampling temperature & top/repair 0.0; action 0.1 (2) & fixed by layer: deterministic state/repair and low-variance action realization \\
Retries/site concurrency & 1/1 (1 each) & fixed to isolate one retry and prevent simultaneous tasks on the same website \\
Workers & 2, 4, 6, 8, 12, 16 (6) & 6--16 by provider/run; throughput only, chosen below the global rate limit \\
Global RPM & 25, 60, 120, 500 (4) & provider-specific quota; throughput only and shared across workers \\
\bottomrule
\end{tabularx}
\caption{Controller and runtime values used during development. Worker and RPM settings affect collection throughput, not the per-contract policy or scoring rule.}
\label{tab:controller-development-app}
\end{appxtablewide}

\paragraph{Run provenance.}
All live Gemini rows use the same frozen 4,096-contract release ledger, whose
SHA-256 is
\path{ac92cdc24c1a2ded383b0bdb8aa3c33711682003781a96bfb55d216e08b73368}.
The anonymous artifact indexes each canonical task-level ledger by the stable
archive ID and evaluator version in \autoref{tab:run-provenance-app}.

\begin{appxtable}
\footnotesize
\begin{tabularx}{\linewidth}{@{}>{\raggedright\arraybackslash}p{1.70cm}>{\raggedright\arraybackslash}X>{\raggedright\arraybackslash}p{1.20cm}@{}}
\toprule
Paper family & Artifact manifest ID & Eval. \\
\midrule
Full-Pro & \texttt{WR-FULL-PRO} & \texttt{CGS-v1} \\
Full-Flash & \texttt{WR-FULL-FLASH} & same \\
Flash-60 & \texttt{WR-FLASH-60} & same \\
Teacher/full & \texttt{WR-TEACHER-1024} & same \\
Structural arms & \texttt{WR-STRUCT-1024} & same \\
IntentCore-30 & \texttt{WR-INTENT-30} & same \\
Exec-Pro & \texttt{WR-EXEC-PRO} & same \\
Exec-Flash & \texttt{WR-EXEC-FLASH} & same \\
\bottomrule
\end{tabularx}
\caption{Paper names mapped to stable task-level artifact IDs and calibrated evaluator versions. Training and external-proxy artifacts are described in \autoref{app:trained-middle} and \autoref{app:fullscale-diagnostics}.}
\label{tab:run-provenance-app}
\end{appxtable}

\paragraph{Repeated collection and matched comparisons.}
Live rollout, evaluation, and ablation jobs were repeated in three interleaved weekly waves. Within a wave, corresponding runs used the same contract file and calibrated audit version; paired mechanism comparisons also used the same contracts and step budget. The paper reports all-contract ledgers, same-contract paired slices, or explicitly labeled clean/nonblocked diagnostics rather than mixing scopes.

\paragraph{Randomness and replay.}
\autoref{tab:randomness-app} consolidates every seed that changes a reported
sample, split, side assignment, or rollout schedule.  The stable runner stores
the seed in its manifest and reuses it on resume; task shuffling, site
round-robin tie breaking, and launch jitter use a local \texttt{Random}
instance derived from that seed.  Hosted model APIs do not expose a portable
generation seed, and public websites are time-varying.  Consequently, the
archived task IDs, outputs, screenshots, manifests, and hashes reproduce the
evaluated sample exactly, while a new live request is not expected to be
bitwise identical.  Distributed training likewise fixes model/data seeds but
does not claim bitwise-deterministic CUDA execution.

\begin{appxtable}
\footnotesize
\begin{tabularx}{\linewidth}{@{}>{\raggedright\arraybackslash}p{1.90cm}p{1.55cm}>{\raggedright\arraybackslash}X@{}}
\toprule
Procedure & Seed & Controlled randomness \\
\midrule
Full Pro/Flash & 20260518 & task order, site interleaving, and launch jitter \\
Structural arms & 20260521 & common 1,024-contract schedule across controller arms \\
Horizon runs & 20260525 & Flash-60 and 30-step diagnostic schedules \\
Exec Pro/Flash & 20260606 & common 4,096-contract executable-boundary schedule \\
GPT diagnostic & 20260612 & blocked-site-excluded task schedule \\
Human packets & 20260520 & stratified packet draw and blinded left/right assignment \\
Model-rater audit & 20260605 & stratified diagnostic subsample; temperature 0 \\
Contract training split & 20260606 & deterministic hash split by \texttt{base\_task\_id} \\
Expanded training split & 20260613 & deterministic hash split by \texttt{base\_task\_id} \\
Optimization & 42 & model initialization path, data order, and trainer sampler \\
External proxies & \makecell[l]{20260610\\20260612} & paired example order for Mind2Web/WebLINX adapters \\
\bottomrule
\end{tabularx}
\caption{Seeds for reported stochastic procedures. API responses and the live web remain externally nondeterministic; evaluated artifacts are immutable and checksum-addressed.}
\label{tab:randomness-app}
\end{appxtable}

\paragraph{Automatic audit.}
Evidence and answer gates combine GPT-5.5 high-reasoning and Gemini 3.1 Pro preview judgments with deterministic checks. The audit is calibrated against two experts on 300 rollouts; \autoref{app:human-audit} reports gate-level agreement and false positives. External-access and runtime failures receive separate detected labels.

\section{Appendix F: Full Live and Mechanism Results}
\refstepcounter{appendixref}
\label{app:fullscale-diagnostics}

This section expands the compact main tables. It first reports paired transitions and external proxy checks, then gives the complete live and structural result rows. Percentages are shown without percent signs inside tables; operational cost remains in \autoref{app:robustness}.

\paragraph{Matched mechanism effects.}
The structural comparison is paired by contract. \autoref{tab:paired-deltas-app} reports exact transitions rather than inferring a paired effect from rounded endpoints. The largest gain from the full hierarchy is evidence satisfaction.

\begin{appxtable}
\footnotesize
\setlength{\tabcolsep}{2.2pt}
\begin{tabularx}{\linewidth}{@{}>{\raggedright\arraybackslash}Xrrrr@{}}
\toprule
Pair / metric & $\Delta$ & Better & Worse & Tie \\
\midrule
Teacher $\rightarrow$ IntentCore / CGS & \textbf{+6.74} & 143 & 74 & 807 \\
Teacher $\rightarrow$ IntentCore / Evid & \textbf{+15.23} & 225 & 69 & 730 \\
Teacher $\rightarrow$ IntentCore / Ans & \textbf{+12.37} & 408 & 190 & 426 \\
Flat $\rightarrow$ Intent-only / CGS & +0.88 & 111 & 102 & 811 \\
Persona-only $\rightarrow$ IntentCore / CGS & \textbf{+4.20} & 135 & 92 & 797 \\
\bottomrule
\end{tabularx}
\caption{Paired structural deltas (percentage points) on the same 1,024 contracts.}
\label{tab:paired-deltas-app}
\end{appxtable}

\paragraph{External mechanism checks.}
External datasets cannot reproduce live-WebRider contracts, but they can test related intention, state, and profile interfaces. \autoref{tab:external-baselines-app} keeps these proxy checks separate from the live leaderboard.

\begin{appxtable}
\footnotesize
\setlength{\tabcolsep}{2.0pt}
\begin{tabular}{@{}lrrrr@{}}
\toprule
Proxy & $N$ & Base & Test & $\Delta$ \\
\midrule
Mind2Web intent & 1,339 & 30.8 & \textbf{42.7} & \textbf{+11.9} \\
WebLINX intent & 500 & 34.8 & \textbf{39.0} & \textbf{+4.2} \\
Mind2Web state & 500 & 36.8 & \textbf{43.6} & \textbf{+6.8} \\
PersonalWAB-style & 120 & 0.0 & \textbf{15.0} & \textbf{+15.0} \\
\bottomrule
\end{tabular}
\caption{External proxy checks. Corresponding $p$ values are $<10^{-12}$, 0.0275, $7.56\times10^{-5}$, and $7.63\times10^{-6}$.}
\label{tab:external-baselines-app}
\end{appxtable}

Mind2Web step success requires the correct operation and target label, plus
token F1 $\geq0.8$ for typed values. WebLINX step success is exact candidate
selection. The PersonalWAB-style metric is exact recovery of the held-out gold
candidate from a fixed set. All significance values in
\autoref{tab:external-baselines-app} use exact paired McNemar tests.
Dataset splits, fixed checkpoints, condition changes, and decoding settings for
these proxy tests are specified in \autoref{tab:external-baseline-setup-app}.

\begin{appxtablewide}
\small
\begin{tabularx}{\linewidth}{@{}>{\raggedright\arraybackslash}p{2.35cm}>{\raggedright\arraybackslash}p{4.10cm}>{\raggedright\arraybackslash}p{4.15cm}X@{}}
\toprule
Proxy & Dataset split & Fixed model & Condition change \\
\midrule
Mind2Web intent & \path{osunlp/Multimodal-Mind2Web} \texttt{test\_task}; 31 negatives & Qwen3-VL-8B + MidSFT-v3; $T=0$ & semantic intention vs. task-only \\
Mind2Web state & first 500 paired Mind2Web rows & Qwen3-VL-8B + MidSFT-v3; same prompt; $T=0$ & screenshot vs. screenshot removed \\
WebLINX intent & \path{McGill-NLP/WebLINX} \path{reranking/validation}; 31 negatives & Qwen3-VL-8B + MidSFT-v3; $T=0$ & full dialogue intention vs. last user utterance \\
PersonalWAB-style & public test split; ambiguous task view; 7 negatives; 8 history items & Gemini 3.5 Flash; $T=0$ & profile+history vs. task-only \\
\bottomrule
\end{tabularx}
\caption{External-proxy setup. Mind2Web and WebLINX use MidSFT-v3 with seeds 20260610/20260612; the PersonalWAB-style test uses Gemini 3.5 Flash with temperature zero.}
\label{tab:external-baseline-setup-app}
\end{appxtablewide}

\paragraph{Live audits.}
\autoref{tab:all-live-results-app} separates the contract-quality components from access and runtime conditions. Full-Pro exposes the gap between termination and contract satisfaction; Flash-60 changes only the horizon; Exec-Pro and Exec-Flash move all finalization out of the middle. GPT-5.5-Diag is a 1,024-contract diagnostic slice that excludes five heavily access-blocked websites.

\begin{appxtable}
\footnotesize
\setlength{\tabcolsep}{1.55pt}
\begin{tabular}{@{}lrrrrrr@{}}
\toprule
Run & $N$ & Term & CGS & Hard & Evid & Ans \\
\midrule
Full-Pro & 4096 & 99.2 & \textbf{38.8} & 70.3 & 57.4 & 42.3 \\
Full-Flash & 4096 & 99.7 & 34.4 & 68.2 & 58.6 & 62.5 \\
Flash-60 & 4096 & 99.0 & 37.9 & \textbf{73.3} & \textbf{58.8} & \textbf{64.1} \\
Exec-Pro & 4096 & 99.7 & 32.9 & 42.2 & 35.7 & 33.0 \\
Exec-Flash & 4096 & 99.5 & 32.8 & 41.3 & 35.6 & 33.6 \\
GPT-5.5-Diag & 1024 & -- & 44.9 & 67.2 & 62.5 & 46.1 \\
\bottomrule
\end{tabular}
\caption{Complete live-result ledger: terminal completion and contract-quality components (\%).}
\label{tab:all-live-results-app}
\end{appxtable}

\autoref{tab:all-live-operations-app} separates access, runtime, and browser-step conditions from contract-quality outcomes.

\begin{appxtable}
\footnotesize
\setlength{\tabcolsep}{3.0pt}
\begin{tabular}{@{}lrrr@{}}
\toprule
Run & Access & Runtime & Mean steps \\
\midrule
Full-Pro & 13.5 & 0.8 & 8.37 \\
Full-Flash & 15.7 & 0.3 & 11.49 \\
Flash-60 & 17.0 & 1.0 & 20.04 \\
Exec-Pro & 14.7 & 0.3 & 9.09 \\
Exec-Flash & 16.9 & 0.5 & 11.58 \\
GPT-5.5-Diag & 3.1 & 0.0 & -- \\
\bottomrule
\end{tabular}
\caption{Access and runtime conditions for the live rows (\%, except mean browser-action steps).}
\label{tab:all-live-operations-app}
\end{appxtable}

\paragraph{Prompt and structural controls.}
All rows in \autoref{tab:all-ablations-app} use the same 1,024-contract slice and 20-step cap except IntentCore-30, the labeled horizon diagnostic. Prompt+PI and IntentCore contain the same persona/intention information; the difference is whether that information is bound to persistent top state and one guarded action interface.

\begin{appxtable}
\footnotesize
\setlength{\tabcolsep}{1.8pt}
\begin{tabular}{@{}lrrrrr@{}}
\toprule
Condition & CGS & Hard & Evid & Ans & Early ans. \\
\midrule
Prompt & 36.5 & 66.0 & 52.0 & 55.0 & 31 \\
Prompt+PI & 41.2 & 70.5 & 60.5 & 62.0 & 24 \\
Teacher/full & 40.0 & 72.6 & 56.8 & 59.3 & -- \\
Flat & 44.6 & 82.4 & 70.9 & 71.1 & $\sim$14 \\
Persona-only & 42.6 & \textbf{84.1} & 70.6 & 70.6 & -- \\
Intent-only & 45.5 & 81.0 & 69.1 & \textbf{71.8} & -- \\
IntentCore & \textbf{46.8} & 78.6 & \textbf{72.1} & 71.6 & $\sim$12 \\
IntentCore-30 & 42.0 & 76.5 & 65.6 & 68.4 & -- \\
\bottomrule
\end{tabular}
\caption{Prompt and structural ablations (\%). The 30-step row is a horizon diagnostic, not an architecture comparison.}
\label{tab:all-ablations-app}
\end{appxtable}

\paragraph{Horizon operation check.}
The matched operation statistics in \autoref{tab:intentcore-horizon-app} show
that the additional ten steps neither rescue IntentCore nor leave the runtime
distribution unchanged: CGS falls 4.8 points while timeout/runtime failures
increase by 3.81 points.

\begin{appxtable}
\footnotesize
\setlength{\tabcolsep}{2.8pt}
\begin{tabular}{@{}lrrrr@{}}
\toprule
Condition & Cap & $N$ & CGS & Runtime / mean steps \\
\midrule
IntentCore & 20 & 1024 & \textbf{46.8} & \textbf{1.17 / 9.21} \\
IntentCore-30 & 30 & 1024 & 42.0 & 4.98 / 11.43 \\
\bottomrule
\end{tabular}
\caption{Matched IntentCore horizon diagnostic (\%, except browser-action steps). Runtime includes timeouts and runtime exceptions in the calibrated task ledger.}
\label{tab:intentcore-horizon-app}
\end{appxtable}

\section{Appendix G: Human Stepwise Evaluation}
\refstepcounter{appendixref}
\label{app:human-audit}

\begin{figure*}[t]
\centering
\includegraphics[width=0.90\linewidth]{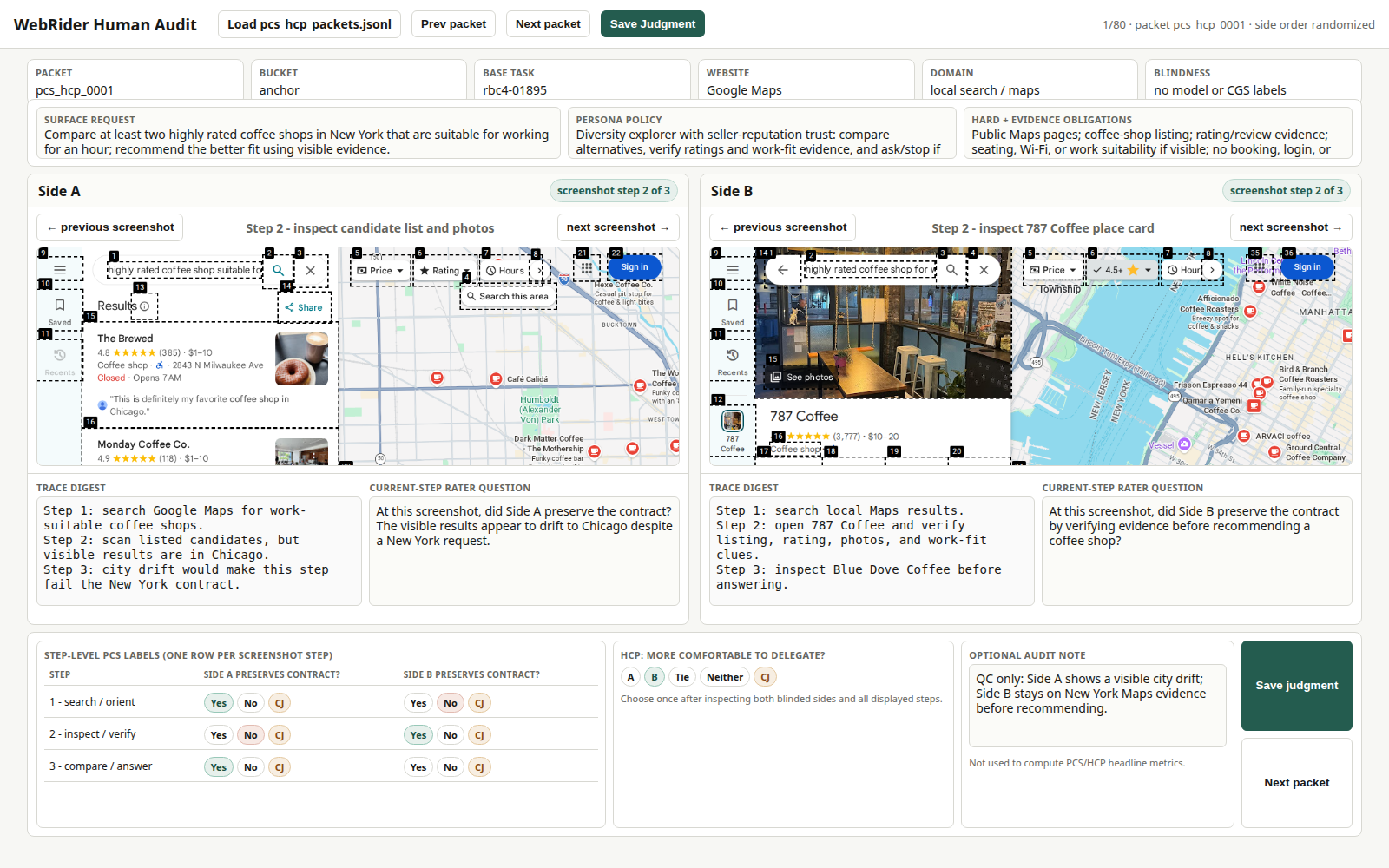}
\caption{Human stepwise-audit interface. The contract remains visible while raters inspect two blinded rollouts, mark whether displayed steps preserve the persona policy, and choose the rollout they would rather delegate to.}
\label{fig:human-audit-interface}
\end{figure*}

The human study asks whether a person would accept the visible route under the delegated policy. Raters see the same contract and two randomized sides, but not run names or automatic CGS labels. \autoref{fig:human-audit-interface} shows the interface; \autoref{tab:human-workflow-app} maps the study design to the reported statistics.
Six adult university-affiliated volunteers were recruited by convenience
sampling: two laboratory colleagues served as expert calibrators and four
other university members served as non-expert raters. Participation was
voluntary and uncompensated; all raters consented before labeling.

\begin{appxtable}
\footnotesize
\begin{tabularx}{\linewidth}{@{}p{1.65cm}X@{}}
\toprule
Stage & Unit and output \\
\midrule
Contract review & common request, policy, constraints, and evidence obligations for both sides \\
Step audit & yes/no/cannot-judge per displayed step; aggregated to rollout-level persona consistency \\
Pair preference & A/B/tie/neither/cannot-judge; aggregated to net human comfort \\
Shared anchors & 20 pairs rated by all six raters; agreement and expert--amateur checks \\
Coverage pairs & 60 pairs, two raters per pair; study coverage \\
Gate calibration & 300 rollouts, two experts; automatic-gate agreement and false positives \\
\bottomrule
\end{tabularx}
\caption{Human-study workflow and statistic mapping.}
\label{tab:human-workflow-app}
\end{appxtable}

\paragraph{Blinding and aggregation.}
The interface assigns conditions to left and right with a seeded random draw
for each packet--rater assignment. Run names, condition identities, and
automatic CGS labels remain hidden until labels are locked. Shared-anchor pairs
receive six independent ratings. Coverage pairs receive two independent
non-expert ratings after rubric calibration. For a step, a strict majority of
judgeable ratings determines yes or no; a yes/no tie, or a step with no
judgeable rating, becomes CJ. Thus a two-rater coverage disagreement is CJ at
the step level. For pair preference, a strict majority determines A, B, or
neither; any remaining disagreement becomes tie. Neither means both routes are
unacceptable, whereas CJ is reserved for missing or illegible information.

\paragraph{PCS.}
For rollout $r$, every displayed step $s$ is first aggregated across its
assigned raters to $\tilde y_{rs}\in\{1,0,\mathrm{CJ}\}$ by the rule above.
Let $C_r=\{s:\tilde y_{rs}=1\}$ and
$J_r=\{s:\tilde y_{rs}\in\{0,1\}\}$:
\[
 \mathrm{PCS}_r=\frac{|C_r|}{|J_r|}.
\]
CJ steps are excluded from the denominator and reported separately; a rollout
with no judgeable step remains CJ. Arm-level PCS pools aggregated consistent
and judgeable steps for description. Wilson intervals accompany these pooled
rates. Confirmatory inference preserves the matched design: it applies a
two-sided Wilcoxon signed-rank test to the 80 rollout-level differences
$\mathrm{PCS}_{A,r}-\mathrm{PCS}_{B,r}$. For example, aggregated labels
\{yes, yes, CJ, no, yes\} give $\mathrm{PCS}=3/4=0.75$; against a matched
rollout at 0.50, the pair difference is $+0.25$.

\paragraph{HCP.}
Let $n_A,n_B,n_T,n_N$ denote the packet-level A, B, tie, and neither counts.
True cannot-judge packets are reported separately and excluded, giving
$N=n_A+n_B+n_T+n_N$. The primary all-pair net preference is
\[
 \Delta_{\mathrm{HCP}}=(n_A-n_B)/N.
\]
This denominator retains ties and neither outcomes. We also report the
decisive A-win rate $n_A/(n_A+n_B)$ with a Wilson interval and test the null of
equal A/B preference using a two-sided exact binomial sign test over the
$n_A+n_B$ decisive pairs. The 80 packets yield
$(n_A,n_T,n_B,n_N)=(47,16,13,4)$ and no packet-level CJ: all-pair net HCP is
$42.5$ points; among 60 decisive pairs, A wins $78.3\%$ (95\% Wilson CI
$[66.4,86.9]$; exact $p=1.21\times10^{-5}$).

Arm A is IntentCore and Arm B is the teacher/full controller. \autoref{tab:human-pilot-headline-app} reports the paired outcome. Anchor agreement is $\kappa=0.66$ for PCS and $\kappa=0.62$ for HCP; expert--non-expert agreement is 85\% ($\kappa=0.64$).

\begin{appxtable}
\footnotesize
\setlength{\tabcolsep}{3.0pt}
\begin{tabularx}{\linewidth}{@{}>{\raggedright\arraybackslash}Xrrrr@{}}
\toprule
Measure & $N$ & A & B & $\Delta$/net \\
\midrule
Persona consistency & 80 pairs & \textbf{578/672} & 469/671 & \textbf{+16.1 pp} \\
Human comfort & 80 pairs & \textbf{47 prefer} & 13 prefer & \textbf{+42.5 pp} \\
\bottomrule
\end{tabularx}
\caption{Human-study outcomes. PCS entries are aggregated consistent/judgeable steps; HCP retains 16 ties and four neither outcomes in its all-pair denominator. A is IntentCore and B is the teacher/full controller.}
\label{tab:human-pilot-headline-app}
\end{appxtable}

Pooled PCS is 86.0\% for A (95\% Wilson CI $[83.2,88.4]$) and 69.9\%
for B ($[66.3,73.2]$); the matched rollout-level Wilcoxon test gives
$p<0.001$. Across the 160 rollouts, persona consistency and CGS correlate at
$\rho\approx0.55$; 11\% of CGS-passing rollouts still receive a low
persona-consistency score. The two evaluations therefore overlap only partly.

The automatic gates are strongest for terminal and external-access detection and noisier for evidence and answer quality. \autoref{tab:auto-expert-calibration-app} reports calibration against two experts on 300 rollouts.

\begin{appxtable}
\footnotesize
\setlength{\tabcolsep}{2.4pt}
\begin{tabular}{@{}lrrr@{}}
\toprule
Gate & Accuracy & $\kappa$ & False positive \\
\midrule
Terminal & \textbf{0.99} & 0.93 & 0.6\% \\
Hard-C & 0.91 & 0.81 & 5.1\% \\
Evidence & 0.86 & 0.72 & 8.4\% \\
Answer & 0.87 & 0.74 & 6.8\% \\
External access & 0.98 & \textbf{0.95} & \textbf{0.4\%} \\
CGS & 0.88 & 0.74 & 6.0\% \\
\bottomrule
\end{tabular}
\caption{Automatic gates against two experts on 300 rollouts. Evidence and answer judgments are the noisiest components.}
\label{tab:auto-expert-calibration-app}
\end{appxtable}

Model judges were also tested as rubric diagnostics. Their abstention and alignment in \autoref{tab:model-judge-pilots-app} explain why these pilots are not treated as human gold.

\begin{appxtable}
\footnotesize
\setlength{\tabcolsep}{2.2pt}
\begin{tabular}{@{}lrrrr@{}}
\toprule
Diagnostic & $N$ & \makecell[r]{PCS\\align.} & \makecell[r]{HCP\\align.} & CJ/adjud. \\
\midrule
ChatGPT packet & 50 & 44\% & 42\% & 8/50 \\
ChatGPT text & 200 & -- & -- & 86.5\% \\
Gemini image-only & 50 & -- & -- & 36/50 \\
Gemini records+image & 50 & -- & -- & 44/50 \\
\bottomrule
\end{tabular}
\caption{Model-rater diagnostics. These rows test rubric sufficiency and are excluded from the PCS/HCP human results.}
\label{tab:model-judge-pilots-app}
\end{appxtable}

\FloatBarrier

\section{Appendix E: Training the Action-Realization Layer}
\refstepcounter{appendixref}
\label{app:trained-middle}

Training the middle directly tests the hierarchy's reusable action boundary. The top layer retains the contract, evidence ledger, and ask/answer/stop decisions; deterministic guards retain the legal action grammar; and the executor remains fixed. The learned component maps a top intention and browser observation to one post-guard executable action. This boundary yields a supervised target at every step while leaving long-horizon finalization with the top layer.

\paragraph{Training records.}
Each record contains the screenshot, task contract, persona policy, top intention/command, protected constraints, recent history, and visible UI labels. The target is the final executable Action AST after compiler repair and deterministic guards. Terminal actions are removed because finalization belongs to the top layer.

The v3 exporter began with 34,297 Full-Pro step records and retained 27,246
high-confidence executable targets. The 7,051 exclusions comprise 3,633
low-confidence rows, 250 rows without an action type, 2,234 terminal actions,
and 934 rows whose top subgoal was terminal. A deterministic hash of
\texttt{base\_task\_id} assigns complete base-task groups to the 25,100/2,146
train/evaluation split (8\%, seed 20260606), preventing sibling persona
contracts from crossing the split. Sites remain represented on both sides;
this is a base-task-grouped split, not a cross-site generalization test.

\autoref{tab:live-middle-app} reports the live replacement under the fixed top/runtime boundary. All three rows use the current 1024-contract scope; clean CGS removes external-access and runtime failures only.

\begin{appxtable}
\footnotesize
\setlength{\tabcolsep}{2.8pt}
\begin{tabular}{@{}lrrrrr@{}}
\toprule
Middle & CGS & Hard & Evid & Access & Clean \\
\midrule
Exec-Pro ref. & 31.2 & -- & -- & 14.7 & 38.9 \\
Exec-Flash ref. & 34.8 & -- & -- & 16.9 & 39.9 \\
MidSFT-8B & \textbf{50.8} & \textbf{61.3} & \textbf{78.1} & \textbf{2.0} & \textbf{52.0} \\
\bottomrule
\end{tabular}
\caption{Live middle replacement (\%) on 1,024 contracts, with the top layer and deterministic guards fixed.}
\label{tab:live-middle-app}
\end{appxtable}

The offline comparison in \autoref{tab:trained-middle-offline-app} evaluates the same post-guard prediction problem. Qwen3-VL-8B v3 gives the best held-out action match among the evaluated students; increasing parameter count does not improve this boundary across the evaluated configurations.

\begin{appxtable}
\footnotesize
\setlength{\tabcolsep}{3.0pt}
\begin{tabular}{@{}lrrr@{}}
\toprule
Student & Eval $N$ & Action & Target \\
\midrule
MidSFT-8B v3 & 2,146 & \textbf{91.24} & \textbf{84.81} \\
Qwen8B-Exec & 2,422 & 88.93 & 82.33 \\
Qwen32B & 1,024 & 90.23 & 82.81 \\
Gemma4-31B & 1,024 & 72.27 & 79.69 \\
Qwen30B-A3B & 128 & 82.81 & 75.78 \\
\bottomrule
\end{tabular}
\caption{Held-out post-guard action and target match (\%). Evaluation sizes follow the completed run for each student.}
\label{tab:trained-middle-offline-app}
\end{appxtable}

AST validity requires a parseable JSON object with a normalized action type and
a fallback object. \emph{Action match} is exact equality of the normalized
predicted and guarded-target action types; \emph{target match} is exact equality
of their visible UI labels (including the empty label for actions without a UI
target). These metrics isolate executable realization and do not score the top
layer's evidence-sufficiency or final-answer decision.

Model identities and corpus sizes are explicit in \autoref{tab:training-models-app}.
The release manifest records the corpus hash, split seed, evaluated adapter
checkpoint, stable artifact ID, and base-task ID for every record.
\begin{appxtable}
\footnotesize
\begin{tabularx}{\linewidth}{@{}l>{\raggedright\arraybackslash}X>{\raggedright\arraybackslash}p{1.95cm}@{}}
\toprule
Student & Base checkpoint & Corpus; train/eval \\
\midrule
MidSFT-8B v3 & \path{Qwen/Qwen3-VL-8B-Instruct} & Contract; 25,100/2,146 \\
Qwen8B-Exec & \path{Qwen/Qwen3-VL-8B-Instruct} & Exec-expanded; 29,170/2,422 \\
Qwen32B & \path{Qwen/Qwen3-VL-32B-Instruct} & Contract; 25,100/2,146 \\
Gemma4-31B & \path{google/gemma-4-31B-it} & Contract; 25,100/2,146 \\
Qwen30B-A3B & \path{Qwen/Qwen3-VL-30B-A3B-Instruct} & Exec-expanded; 29,170/2,422 \\
\bottomrule
\end{tabularx}
\caption{Base checkpoints and executable-only corpora. The Contract split uses seed 20260606; the expanded executable split uses seed 20260613.}
\label{tab:training-models-app}
\end{appxtable}

\autoref{tab:training-setup-app} records the common optimization recipe;
\autoref{tab:training-hardware-app} gives model-specific hardware and distributed
settings recovered from the final checkpoint metadata. All runs use the same strict filtering rule and optimizer seed.

\begin{appxtablewide}
\small
\begin{tabularx}{\linewidth}{@{}>{\raggedright\arraybackslash}p{2.25cm}>{\raggedright\arraybackslash}p{4.05cm}>{\raggedright\arraybackslash}p{4.10cm}X@{}}
\toprule
Aspect & Core setting & Key values & Fixed detail \\
\midrule
Target & post-guard Action AST & supervised executable action & terminal decisions excluded \\
Adapter & bf16 PEFT LoRA & $r=16$; $\alpha=32$; dropout 0.05 & no bias or saved dense modules; no quantization \\
Trainable modules & dense: language $q/k/v/o$ + MLP gate/up/down & MoE: language $q/k/v/o$ & vision tower and multimodal aligner frozen \\
Input / precision & bf16; gradient checkpointing & length 6,144; RGB screenshots & processor defaults; no pixel or image-token cap \\
Optimizer & fused AdamW & $\beta=(0.9,0.95)$; $\epsilon=10^{-8}$; decay 0.1 & gradient norm 1.0 \\
Schedule & cosine; zero warmup & one corpus pass; seeds 42 & fixed across reported runs \\
Data pipeline & shuffled; lazy; right-padded & no packing or drop-last & delete overlength; one preprocessing worker \\
Checkpointing & final-step selection & eval/250; W\&B log/5; retain 3 & load-best-at-end disabled \\
\bottomrule
\end{tabularx}
\caption{Common middle-policy optimization settings.}
\label{tab:training-setup-app}
\end{appxtablewide}

\paragraph{Development range and selection.}
\autoref{tab:training-development-app} enumerates the distinct values in the
archived launch configurations and smoke logs.  These were sequential
fit/stability checks rather than a full Cartesian search.  Final settings were
chosen without live-test CGS: first by successful full-corpus execution within
the four-A6000 memory envelope, then by held-out executable-action metrics.
MidSFT-8B is the principal student because it has the highest held-out action
match among the completed students under the corrected middle-only target.
Every remaining optimization field in \autoref{tab:training-setup-app} was
held at its one reported value and was not tuned.

\begin{appxtablewide}
\small
\begin{tabularx}{\linewidth}{@{}>{\raggedright\arraybackslash}p{2.35cm}>{\raggedright\arraybackslash}p{4.15cm}>{\raggedright\arraybackslash}p{4.10cm}X@{}}
\toprule
Parameter & Values tried (count) & Selected & Criterion \\
\midrule
Adaptation & 4-bit QLoRA; bf16 LoRA (2) & bf16 LoRA & full-corpus fit on 4$\times$A6000; no quantization \\
LoRA $r/\alpha$ & 8/16; 16/32 (2) & 16/32 & rank 16 fit the full runs \\
Learning rate & $10^{-4}$; $8{\times}10^{-5}$; $6{\times}10^{-5}$; $3{\times}10^{-5}$; $2{\times}10^{-5}$; $1.5{\times}10^{-5}$; $10^{-5}$ (7) & \makecell[l]{$8{\times}10^{-5}$ (8B)\\$3{\times}10^{-5}$ (32B)\\$2{\times}10^{-5}$ (MoE)\\$10^{-5}$ (Gemma recovery)} & stable completion; held-out action match \\
Maximum length & 3072; 4096; 6144; 8192 (4) & 6144 & longest validated setting across dense runs \\
Microbatch & 1; 2; 3 (3) & 1 (dense); 2 (MoE) & largest value with memory headroom \\
Gradient accumulation & 1; 2; 4; 8; 11; 16 (6) & per model in \autoref{tab:training-hardware-app} & effective global batch 32 (33 for Gemma recovery) \\
Epochs & 1 (1; fixed) & 1 & one complete corpus pass \\
LoRA dropout & 0.05 (1; fixed) & 0.05 & fixed across full runs \\
Optimizer / schedule & AdamW; cosine (1 each) & AdamW; cosine & fixed $\beta$, $\epsilon$, decay, and zero warmup \\
\bottomrule
\end{tabularx}
\caption{Middle-policy development values and final-selection criteria. Smoke-only values are counted because they informed feasibility; final per-model settings appear in \autoref{tab:training-hardware-app}.}
\label{tab:training-development-app}
\end{appxtablewide}

\begin{appxtablewide}
\small
\setlength{\tabcolsep}{4.0pt}
\begin{tabular}{@{}lrrrrrrr@{}}
\toprule
Run & GPUs & Micro & Accum. & Global & Learning rate & ZeRO & Final step \\
\midrule
MidSFT-8B & 4 & 1 & 8 & 32 & $8\times10^{-5}$ & Z2 & 785 \\
Qwen8B-Exec & 4 & 1 & 8 & 32 & $8\times10^{-5}$ & Z2 & 912 \\
Qwen32B & 4 & 1 & 8 & 32 & $3\times10^{-5}$ & Z3 & 785 \\
Gemma4-31B & $4\to3$ & 1 & $8\to11$ & $32\to33$ & $3\times10^{-5}\to10^{-5}$ & Z3 & 785$^*$ \\
Qwen30B-A3B & 4 & 2 & 4 & 32 & $2\times10^{-5}$ & Z3 & 912 \\
\bottomrule
\end{tabular}
\captionsetup{width=0.78\linewidth}
\caption{Model-specific settings. GPUs are 48-GiB RTX A6000s; Global is GPUs $\times$ microbatch $\times$ accumulation.}
\label{tab:training-hardware-app}
\end{appxtablewide}

All runs used step-based evaluation and checkpointing. The 8B and MoE runs
saved every 250 steps; Qwen32B and Gemma saved every 100. For Gemma, a change
from four to three workers made the four-rank optimizer state incompatible;
the step-600 adapter weights were restored model-only, and AdamW and its
schedule were reinitialized for steps 601--785 (the asterisk in the table).

\paragraph{Software environment.}
Training used the recorded environment in \autoref{tab:training-software-app}.
Live collection was CPU-only and requested 32 vCPUs and 128\,GiB RAM per
managed worker job.  The scheduler exposed this logical allocation but not the
physical host CPU SKU; hosted Gemini/OpenAI serving hardware was likewise
provider-managed.  \autoref{tab:runtime-software-app} therefore reports the
complete reproducible client environment and explicitly marks the two
provider-controlled quantities.

\begin{appxtablewide}
\small
\begin{tabularx}{0.84\linewidth}{@{}p{2.60cm}X@{}}
\toprule
Component & Recorded environment \\
\midrule
Host & Ubuntu 24.04 LTS; Linux 6.8.0-124-generic (x86-64, glibc 2.39); AMD Ryzen Threadripper 3990X; 251\,GiB RAM \\
Accelerators & 4$\times$NVIDIA RTX A6000, 48\,GiB each; driver 570.207 \\
Core stack & Python 3.12.13; PyTorch 2.11.0+cu128; CUDA 12.8 \\
Training stack & Transformers 5.8.1; Accelerate 1.13.0; DeepSpeed 0.19.1; ms-swift 4.2.3 \\
\bottomrule
\end{tabularx}
\captionsetup{width=0.84\linewidth}
\caption{Recorded software and hardware environment for middle-policy training.}
\label{tab:training-software-app}
\end{appxtablewide}

\begin{appxtablewide}
\small
\begin{tabularx}{0.88\linewidth}{@{}p{1.80cm}X@{}}
\toprule
Component & Recorded live-runtime environment \\
\midrule
Allocation & managed Linux CPU job; 32 vCPUs, 128\,GiB RAM; physical CPU SKU and host kernel not exposed by scheduler \\
Base system & x86-64 Debian GNU/Linux 13.4 (trixie) container userland; Python 3.12.13 \\
Browser & Chromium 148.0.7778.167 and matching ChromeDriver 148.0.7778.167 \\
Full image & google-genai 2.4.0, Selenium 4.44.0, CloakBrowser 0.3.28, NumPy 2.4.5, Pillow 12.2.0 \\
Exec image & google-genai 2.8.0, Selenium 4.44.0, CloakBrowser 0.3.31, NumPy 2.4.6, Pillow 12.2.0 \\
Common utility & python-dotenv 1.2.2; headless Chromium; 1024$\times$768 viewport \\
Hosted models & exact model IDs are in \autoref{tab:run-names-app}; server CPU/GPU and model-build hashes are not exposed by the API providers \\
\bottomrule
\end{tabularx}
\captionsetup{width=0.88\linewidth}
\caption{Hardware allocation, operating system, browser, and client-library versions for live experiments.}
\label{tab:runtime-software-app}
\end{appxtablewide}

\section{Appendix H: Prompt and Executable-Action Interfaces}
\refstepcounter{appendixref}
\label{app:runtime-prompts}

The reproducible object is the interface between layers, not a collection of undocumented prompts. \autoref{fig:prompt-cards-app} gives condensed prompt cards, while \autoref{tab:prompt-contracts-app} states the corresponding input/output contracts. The release contains the full prompt files.

\begin{figure*}[t]
\centering
\begin{minipage}[t]{0.31\linewidth}
\textbf{Top state}\par\smallskip
\begin{lstlisting}[style=webriderprompt]
INPUT
  contract, persona
  observation, ledger
  last action
OUTPUT JSON
  intention
  active constraints
  missing evidence
  blockers
  decision: browse | ask |
            answer | stop
RULE
  never choose a UI label
\end{lstlisting}
\end{minipage}\hfill
\begin{minipage}[t]{0.34\linewidth}
\textbf{Executable middle}\par\smallskip
\begin{lstlisting}[style=webriderprompt]
INPUT
  screenshot, top command
  protected constraints
  recent history, UI labels
OUTPUT JSON
  thought: short grounded note
  action: exactly one of
    click | type | scroll |
    search | maps | back | wait
  fallback: executable action
RULE
  no answer/ask/stop labels
  never invent target labels
\end{lstlisting}
\end{minipage}\hfill
\begin{minipage}[t]{0.31\linewidth}
\textbf{Middle SFT record}\par\smallskip
\begin{lstlisting}[style=webriderprompt]
USER RECORD
  screenshot
  task contract
  persona policy
  top intention/command
  observation and history
ASSISTANT TARGET
  final post-guard
  executable Action AST
FILTER
  remove terminal actions
  keep deterministic fallback
\end{lstlisting}
\end{minipage}
\caption{Condensed prompt interfaces. The training target is the action after deterministic compilation and guards, not raw model text.}
\label{fig:prompt-cards-app}
\end{figure*}

\begin{appxtable}
\footnotesize
\begin{tabularx}{\linewidth}{@{}p{1.65cm}X@{}}
\toprule
Component & Input $\rightarrow$ required output \\
\midrule
Top state & contract, policy, observation, ledger, last action $\rightarrow$ intention, active constraints, missing evidence, blockers, and browse/ask/answer/stop decision \\
Middle & screenshot, top command, protected constraints, recent actions, visible labels $\rightarrow$ one executable Action AST and safe fallback \\
Guard stack & raw AST, repeated state, terminal and external-access signals $\rightarrow$ parsed action, deterministic repair/rejection, recovery, or external-access label \\
Executor & parsed browser/Search/Maps action $\rightarrow$ screenshot, URL/title, visible labels, observation, and status \\
\bottomrule
\end{tabularx}
\caption{Runtime layer contracts. Finalization remains in the top layer; the middle is supervised only on executable action realization.}
\label{tab:prompt-contracts-app}
\end{appxtable}

The structural ablation changes which state fields reach the middle while holding the system prompt, contract slice, model, and step budget fixed. \autoref{tab:ablation-prompt-masks-app} lists the exact state masks, and \autoref{tab:ablation-authority-app} records the held-fixed guard and terminal authority. All arms receive the current observation and a task-local top command. The bounded recent state contains at most five URLs/actions, three failures, loop counts, and the evidence ledger; no arm receives the full raw transcript.

\begin{appxtable}
\footnotesize
\setlength{\tabcolsep}{1.7pt}
\begin{tabular}{@{}lccccc@{}}
\toprule
Mask & Persona & Intent & Ledger & Recent & Raw hist. \\
\midrule
Teacher/full & yes & yes & yes & yes & no \\
Flat & no & no & no & no & no \\
Persona-only & yes & no & no & no & no \\
Intent-only & no & yes & no & no & no \\
IntentCore & yes & yes & no & no & no \\
Exec/MidSFT & yes & yes & yes & yes & no \\
\bottomrule
\end{tabular}
\caption{Actual state fields delivered to the middle. Intent is the structured current intent state; Recent is bounded action/failure history rather than the full trajectory.}
\label{tab:ablation-prompt-masks-app}
\end{appxtable}

\begin{appxtable}
\footnotesize
\begin{tabularx}{\linewidth}{@{}>{\raggedright\arraybackslash}p{2.05cm}>{\raggedright\arraybackslash}p{1.25cm}>{\raggedright\arraybackslash}X@{}}
\toprule
Family & \makecell[l]{Guards/\\compiler} & Terminal authority \\
\midrule
Full/structural & always active & middle proposal accepted or rejected by the terminal gate \\
Exec/MidSFT & always active & top/runtime only; middle grammar is executable-only \\
\bottomrule
\end{tabularx}
\caption{Guard and finalization authority held fixed within each comparison family.}
\label{tab:ablation-authority-app}
\end{appxtable}

The executable grammar remains active for both API and learned middles. \autoref{tab:action-ast-contract-app} defines the legal actions; \autoref{tab:fallback-contract-app} records the bounded recovery policy.

\begin{appxtable}
\footnotesize
\begin{tabularx}{\linewidth}{@{}p{1.25cm}X@{}}
\toprule
Action & Required fields and guarded behavior \\
\midrule
\texttt{click} & visible integer target label; reject missing or prose-only labels \\
\texttt{type} & visible label and nonempty constraint-preserving text \\
\texttt{scroll} & window or visible target plus direction; reject repeated low-information loops \\
\texttt{search} & nonempty query preserving hard constraints; return if the search is unhelpful \\
\texttt{maps} & place query and task location when supplied; legal only for geographic/place tasks \\
Recovery & \texttt{wait} or \texttt{back} with a bounded reason grounded in loading, stale state, access interruption, or browser history \\
Terminal labels & illegal in the executable middle; the top/runtime gate decides answer, incomplete, blocked, or stop \\
\bottomrule
\end{tabularx}
\caption{Action-AST compiler contract. The grammar prevents the middle from taking over evidence sufficiency or finalization.}
\label{tab:action-ast-contract-app}
\end{appxtable}

\begin{appxtable}
\footnotesize
\begin{tabularx}{\linewidth}{@{}p{1.65cm}X@{}}
\toprule
Condition & Allowed response \\
\midrule
Target absent & scroll, search, maps, back, or wait according to top intent \\
Ambiguous target & act only when the top command disambiguates; otherwise observe or search \\
Evidence missing & continue browsing; the middle cannot answer \\
Repeated failure & backtrack, wait, or simplify the query under a guard label \\
External access & stop automation and emit an access-failure signal; do not bypass access policy \\
\bottomrule
\end{tabularx}
\caption{Fallback semantics for executable middle actions.}
\label{tab:fallback-contract-app}
\end{appxtable}

\paragraph{Search and Maps grounding.}
Google Search and Google Maps use the same guarded Action AST as browser-local actions. They are available when page-local evidence is missing, stale, access-blocked, or inherently geographic; retrieved results must still become visible evidence before the top layer may answer. \autoref{tab:grounding-skill-counts-app} reports canonical action-ledger use, and \autoref{fig:grounding-action-template} gives the executable templates.

\begin{appxtable}
\footnotesize
\setlength{\tabcolsep}{2.1pt}
\begin{tabular}{@{}lrrrr@{}}
\toprule
Run & Rows & Search & Maps & Share \\
\midrule
Full-Pro & 34,297 & 2,535 & 9 & 7.42\% \\
Full-Flash & 47,071 & 5,886 & 225 & 12.98\% \\
Flash-60 & 82,087 & 12,311 & 289 & 15.35\% \\
Exec-Pro & 37,252 & 2,165 & 61 & 5.98\% \\
Exec-Flash & 47,418 & 3,895 & 418 & 9.10\% \\
\bottomrule
\end{tabular}
\caption{Grounding-skill use in canonical rollout attempts. Share is Search plus Maps actions divided by action rows.}
\label{tab:grounding-skill-counts-app}
\end{appxtable}

\begin{appxfigure}
\begin{lstlisting}[style=webriderprompt]
SEARCH
{"action":{"type":"google_search",
 "query":"site:<domain> <constraints>"},
 "fallback":{"type":"go_back"}}

MAPS
{"action":{"type":"google_maps",
 "query":"<place/service>",
 "location":"<task city/region>"},
 "fallback":{"type":"google_search",
              "query":"<same constraints>"}}
\end{lstlisting}
\caption{Search and Maps Action-AST templates. Queries preserve hard constraints and persona-relevant preferences; fallbacks return to public evidence without bypassing access controls.}
\label{fig:grounding-action-template}
\end{appxfigure}

\paragraph{Bounded helper decisions.}
The hybrid teacher also exposes the explicit gate outputs in \autoref{tab:helper-prompt-contracts-app}; these helpers never execute browser actions or finalize an answer.

\begin{appxtable}
\footnotesize
\begin{tabularx}{\linewidth}{@{}p{1.35cm}X@{}}
\toprule
Helper & Required output \\
\midrule
Clarifier & \texttt{ASK} only when a missing user fact changes correct behavior; otherwise \texttt{CONTINUE} \\
Verifier & \texttt{ALLOW}, \texttt{CONTINUE}, \texttt{ASK}, or \texttt{BLOCK}, plus missing evidence and violated constraints \\
Recovery & \texttt{OK}, \texttt{LOOP}, \texttt{DEAD\_END}, or \texttt{STALE}, plus a bounded \texttt{BACK}, \texttt{NEW\_SEARCH}, or \texttt{SCROLL} bias \\
\bottomrule
\end{tabularx}
\caption{Teacher-side helper prompt outputs. Deterministic implementations use the same decision vocabulary.}
\label{tab:helper-prompt-contracts-app}
\end{appxtable}

\section{Appendix I: Robustness, Access Effects, and Resources}
\refstepcounter{appendixref}
\label{app:robustness}

This section collects diagnostics needed to interpret the live results without mixing them into the controller comparison. It reports uncertainty for the central estimates, separates external access failures from contract failures, and then gives the pre-specified task, domain, website, and resource views. All percentages use the denominator stated in the caption.

\paragraph{Statistical decision protocol.}
The unit of inference is the contract (or the blinded contract pair for the
human study), and all tests are two-sided at $\alpha=0.05$.  Binary outcomes on
the same contracts use exact McNemar tests, equivalently exact binomial tests
on discordant pairs.  Single proportions use Wilson intervals; paired CGS
effects use the mean of per-contract differences with a 95\% paired interval.
Human PCS uses a paired Wilcoxon signed-rank test over rollout-level PCS
differences, and HCP uses an exact sign test over decisive A/B preferences.
Holm correction is applied within the four archived structural contrasts and
within the four external-proxy contrasts.  The resulting tests and intervals
are consolidated in \autoref{tab:significance-policy-app}. Horizon,
access-adjusted, stratified, and unequal-evaluation-size model rows are
explicitly descriptive; they are not interpreted through unreported
significance tests.

\begin{appxtablewide}
\small
\begin{tabularx}{0.78\linewidth}{@{}>{\raggedright\arraybackslash}p{4.20cm}>{\raggedright\arraybackslash}p{3.40cm}X@{}}
\toprule
Comparison & Test / interval & Result \\
\midrule
Human PCS, A vs. B & paired Wilcoxon & $p<0.001$ \\
Human HCP, A vs. B & exact sign test & 47/60 decisive A wins; $p=1.21{\times}10^{-5}$ \\
Prompt+PI $\rightarrow$ IntentCore & paired 95\% interval & $+5.6$ pp, $[+1.3,+9.9]$; excludes zero \\
Teacher $\rightarrow$ IntentCore & exact McNemar + Holm & 143/74 discordant; $p=3.26{\times}10^{-6}$, $q=1.31{\times}10^{-5}$ \\
Persona-only $\rightarrow$ IntentCore & exact McNemar + Holm & 135/92; $p=0.00520$, $q=0.0156$ \\
Flat $\rightarrow$ IntentCore & exact McNemar + Holm & 126/104; $p=0.166$, $q=0.332$ \\
Flat $\rightarrow$ Intent-only & exact McNemar + Holm & 111/102; $p=0.584$, $q=0.584$ \\
Exec-Pro vs. Exec-Flash & exact McNemar & 299/303; $p=0.903$ \\
Full-Pro $\rightarrow$ Exec-Pro & exact McNemar & 274/516; $p=5.51{\times}10^{-18}$ \\
Full-Flash $\rightarrow$ Flash-60 & exact McNemar & 430/290; $p=2.04{\times}10^{-7}$ \\
External proxies (4) & exact McNemar + Holm & \makecell[l]{corrected $q<4{\times}10^{-12}, 0.0275$;\\$1.51{\times}10^{-4}, 2.29{\times}10^{-5}$} \\
\bottomrule
\end{tabularx}
\captionsetup{width=0.78\linewidth}
\caption{Inferential procedures and reported decisions. Discordant counts are favorable/unfavorable in the arrow direction. Descriptive diagnostics are identified in their table captions.}
\label{tab:significance-policy-app}
\end{appxtablewide}

\paragraph{Uncertainty and matched comparisons.}
\autoref{tab:robustness-app} reports confidence intervals only where a calibrated estimate or archived matched comparison is available. Architecture comparisons use the same contract slice and step budget; Flash-60 is listed separately as a horizon diagnostic.

\begin{appxtable}
\footnotesize
\setlength{\tabcolsep}{1.5pt}
\begin{tabularx}{\linewidth}{@{}>{\raggedright\arraybackslash}Xrr>{\raggedright\arraybackslash}p{1.5cm}@{}}
\toprule
Estimate / comparison & Effect & 95\% CI & Design / $p$ \\
\midrule
Full-Pro CGS & 38.8 & 37.4--40.3 & all contracts \\
Teacher $\rightarrow$ IntentCore & +6.7 & +3.9--+9.5 & matched slice \\
Prompt+PI $\rightarrow$ IntentCore & +5.6 & +1.3--+9.9 & matched slice \\
Flat $\rightarrow$ IntentCore & +2.1 & $-0.8$--+5.0 & matched slice \\
Exec-Pro vs. Exec-Flash & $-0.10$ & $-1.27$--+1.08 & $p=0.903$ \\
Full-Pro $\rightarrow$ Exec-Pro & $-5.91$ & $-7.24$--$-4.58$ & $p=5.5{\times}10^{-18}$ \\
Full-Flash $\rightarrow$ Flash-60 & +3.4 & -- & horizon diagnostic \\
\bottomrule
\end{tabularx}
\caption{Uncertainty for the headline and matched CGS comparisons. Effects are percentage points except the Full-Pro estimate.}
\label{tab:robustness-app}
\end{appxtable}

\paragraph{External access and the denominator.}
External-access failures are detected by the runtime and reported outside the contract gates. \autoref{tab:block-adjusted-app} retains the all-contract result and adds two diagnostics: Unblocked removes contracts with external-access failures; Clean also removes runtime failures. Together, the three denominators expose the access imbalance in the learned-middle comparison.

\begin{appxtable}
\footnotesize
\setlength{\tabcolsep}{2.4pt}
\begin{tabular}{@{}lrrrr@{}}
\toprule
Run & All & Access & Unblocked & Clean \\
\midrule
Exec-Pro & 32.9 & 14.7 & 38.8 & 38.9 \\
Exec-Flash & 32.8 & 16.9 & 39.6 & 39.9 \\
MidSFT-8B & \textbf{50.8} & \textbf{2.0} & \textbf{52.0} & \textbf{52.0} \\
\bottomrule
\end{tabular}
\caption{Access-adjusted CGS (\%). All uses every contract; Unblocked removes external-access failures; Clean also removes runtime failures.}
\label{tab:block-adjusted-app}
\end{appxtable}

\paragraph{Difficulty and ambiguity.}
The author difficulty tags are balanced by construction, while ambiguity is balanced to within one contract. \autoref{tab:difficulty-stratified-app} shows much less variation across these strata than across websites or domains, indicating that access and page affordance drive a large part of live-web difficulty.

\begin{appxtable}
\footnotesize
\setlength{\tabcolsep}{1.5pt}
\begin{tabular}{@{}llrrrr@{}}
\toprule
Slice & Metric & 1/high & 2/med. & 3/low & 4 \\
\midrule
Difficulty & CGS & 39.4 & 40.3 & 38.4 & 37.3 \\
Difficulty & Evid & 58.6 & 56.9 & 59.8 & 54.3 \\
Difficulty & Access & 13.6 & 11.9 & 13.6 & 14.7 \\
\midrule
Ambiguity & CGS & 36.7 & 40.1 & 39.7 & -- \\
Ambiguity & Evid & 56.9 & 57.2 & 58.1 & -- \\
Ambiguity & Access & 13.4 & 13.8 & 13.1 & -- \\
\bottomrule
\end{tabular}
\caption{Full-Pro stratification by difficulty levels 1--4 and ambiguity high/medium/low (\%).}
\label{tab:difficulty-stratified-app}
\end{appxtable}

\paragraph{Domain and task form.}
\autoref{tab:domain-stratified-app} retains all 12 domains but only the metrics needed to explain the spread. \autoref{tab:task-type-stratified-app} shows that task modes have similar access-failure rates but different CGS and evidence satisfaction, so comparison, selection, and stopping structure matter even when access is comparable.

\begin{appxtable}
\footnotesize
\setlength{\tabcolsep}{2.0pt}
\begin{tabular}{@{}lrrr@{}}
\toprule
Domain & CGS & Evid & Access \\
\midrule
ML models/benchmarks & 68.8 & 71.5 & 0.3 \\
Local food/services & 68.5 & 81.8 & 8.2 \\
Open-source tools & 58.5 & 71.5 & 5.6 \\
Electronics/PC retail & 47.6 & 65.6 & 0.3 \\
Everyday shopping & 42.7 & 69.9 & 30.6 \\
Beauty/personal care & 32.6 & 52.9 & 0.0 \\
Scholarly evidence & 32.6 & 55.3 & 1.8 \\
Travel/recreation & 30.6 & 47.6 & 22.6 \\
Home/rental living & 29.4 & 47.9 & 26.2 \\
Recipe/meal planning & 23.8 & 48.2 & 0.0 \\
Public archives & 21.2 & 58.2 & 30.6 \\
Developer docs & 9.4 & 17.6 & 34.4 \\
\bottomrule
\end{tabular}
\caption{Full-Pro results by domain (\%). Each domain contains 340 contracts except everyday shopping (356).}
\label{tab:domain-stratified-app}
\end{appxtable}

\begin{appxtable}
\footnotesize
\setlength{\tabcolsep}{1.9pt}
\begin{tabular}{@{}lrrrr@{}}
\toprule
Task form & $N$ & CGS & Evid & Access \\
\midrule
Compare & 909 & 46.6 & 62.2 & 14.9 \\
Plan & 911 & 42.0 & 58.9 & 12.7 \\
Verify & 664 & 37.3 & 60.1 & 13.0 \\
Lookup & 651 & 34.1 & 53.1 & 13.2 \\
Select & 651 & 30.4 & 52.7 & 13.1 \\
Troubleshoot & 310 & 37.4 & 51.9 & 13.9 \\
\midrule
Evidence answer & 1625 & 36.1 & 55.8 & 13.2 \\
Recommendation & 2471 & 40.7 & 58.5 & 13.6 \\
\bottomrule
\end{tabular}
\caption{Full-Pro results by task mode and output family (\%). Similar access-failure rates isolate differences in evidence collection and stopping.}
\label{tab:task-type-stratified-app}
\end{appxtable}

\autoref{tab:task-mode-difficulty-app} crosses task mode with the balanced difficulty tags. The interaction explains why the marginal difficulty rows are flat: evidence availability and answer form can dominate the author-assigned level.

\begin{appxtable}
\footnotesize
\textit{A. Counts}\par\smallskip
\begin{tabular}{@{}lrrrr@{}}
\toprule
Task form & D1 & D2 & D3 & D4 \\
\midrule
Compare & 243 & 212 & 228 & 226 \\
Plan & 212 & 238 & 226 & 235 \\
Verify & 166 & 172 & 163 & 163 \\
Lookup & 155 & 166 & 172 & 158 \\
Select & 163 & 160 & 161 & 167 \\
Troubleshoot & 85 & 76 & 74 & 75 \\
\bottomrule
\end{tabular}
\par\smallskip
\textit{B. CGS (\%)}\par\smallskip
\begin{tabular}{@{}lrrrr@{}}
\toprule
Task form & D1 & D2 & D3 & D4 \\
\midrule
Compare & 50.2 & 47.6 & 39.9 & 48.7 \\
Plan & 47.6 & 42.0 & 45.6 & 33.6 \\
Verify & 36.1 & 37.8 & 38.0 & 37.4 \\
Lookup & 23.9 & 32.5 & 40.7 & 38.6 \\
Select & 32.5 & 38.8 & 24.2 & 26.3 \\
Troubleshoot & 35.3 & 40.8 & 37.8 & 36.0 \\
\bottomrule
\end{tabular}
\par\smallskip
\textit{C. Evidence (\%)}\par\smallskip
\begin{tabular}{@{}lrrrr@{}}
\toprule
Task form & D1 & D2 & D3 & D4 \\
\midrule
Compare & 63.0 & 62.3 & 61.4 & 61.9 \\
Plan & 65.6 & 57.6 & 61.9 & 51.5 \\
Verify & 62.0 & 57.6 & 62.6 & 58.3 \\
Lookup & 51.6 & 45.8 & 58.1 & 57.0 \\
Select & 51.5 & 60.6 & 58.4 & 40.7 \\
Troubleshoot & 48.2 & 55.3 & 48.6 & 56.0 \\
\bottomrule
\end{tabular}
\par\smallskip
\textit{D. External access (\%)}\par\smallskip
\begin{tabular}{@{}lrrrr@{}}
\toprule
Task form & D1 & D2 & D3 & D4 \\
\midrule
Compare & 14.4 & 16.5 & 12.7 & 15.9 \\
Plan & 13.7 & 9.7 & 11.9 & 15.7 \\
Verify & 12.7 & 12.8 & 14.7 & 11.7 \\
Lookup & 15.5 & 8.4 & 14.5 & 14.6 \\
Select & 11.7 & 10.6 & 14.9 & 15.0 \\
Troubleshoot & 12.9 & 14.5 & 13.5 & 14.7 \\
\bottomrule
\end{tabular}
\caption{Full-Pro task-form by difficulty matrix (\%, except $N$). Easy lookup tasks can be harder than difficult compare tasks because task form changes evidence availability and stopping.}
\label{tab:task-mode-difficulty-app}
\end{appxtable}

\paragraph{Website access.}
\autoref{tab:site-blocked-app} and \autoref{tab:site-zero-block-app} report all 42 websites, sorted by external-access failure rate. The spread shows that live-site access and page affordance explain substantially more variance than the balanced task-difficulty tags.

\begin{appxtable}
\footnotesize
\setlength{\tabcolsep}{2.5pt}
\begin{tabular}{@{}lrrr@{}}
\toprule
Website & $N$ & Access & CGS \\
\midrule
Costco & 84 & 100.0 & 0.0 \\
Stack Overflow & 105 & 100.0 & 0.0 \\
Library of Congress & 105 & 93.3 & 6.7 \\
Wayfair & 96 & 91.7 & 0.0 \\
Expedia & 84 & 89.3 & 4.8 \\
Walmart & 80 & 27.5 & 37.5 \\
Uber Eats & 105 & 24.8 & 50.5 \\
npm & 80 & 21.2 & 71.2 \\
MDN Web Docs & 130 & 9.2 & 14.6 \\
Smithsonian & 105 & 3.8 & 29.5 \\
Amazon & 96 & 3.1 & 71.9 \\
Semantic Scholar & 105 & 2.9 & 26.7 \\
GitHub & 96 & 2.1 & 36.5 \\
arXiv & 130 & 1.5 & 46.9 \\
Wikipedia & 130 & 1.5 & 26.2 \\
Yelp & 130 & 1.5 & 84.6 \\
TripAdvisor & 80 & 1.2 & 38.8 \\
Home Depot & 84 & 1.2 & 35.7 \\
Apple & 96 & 1.0 & 61.5 \\
Booking.com & 96 & 1.0 & 53.1 \\
Papers with Code & 105 & 0.9 & 81.0 \\
PubMed & 105 & 0.9 & 20.9 \\
\bottomrule
\end{tabular}
\caption{Full-Pro website results with nonzero external-access failure rates (\%).}
\label{tab:site-blocked-app}
\end{appxtable}

\begin{appxtable}
\footnotesize
\setlength{\tabcolsep}{2.5pt}
\begin{tabular}{@{}lrrr@{}}
\toprule
Website & $N$ & Access & CGS \\
\midrule
GeeksforGeeks & 105 & 0.0 & 12.4 \\
Google Maps & 105 & 0.0 & 66.7 \\
Hugging Face & 130 & 0.0 & 62.3 \\
IKEA & 80 & 0.0 & 22.5 \\
Kaggle & 105 & 0.0 & 64.8 \\
Micro Center & 80 & 0.0 & 26.2 \\
National Park Service & 80 & 0.0 & 22.5 \\
Newegg & 80 & 0.0 & 78.8 \\
Allrecipes & 96 & 0.0 & 11.5 \\
Nordstrom & 105 & 0.0 & 18.1 \\
Best Buy & 84 & 0.0 & 22.6 \\
PyPI & 84 & 0.0 & 90.5 \\
Docker Hub & 80 & 0.0 & 38.8 \\
Sephora & 130 & 0.0 & 36.9 \\
Epicurious & 80 & 0.0 & 17.5 \\
Serious Eats & 80 & 0.0 & 28.7 \\
Etsy & 80 & 0.0 & 65.0 \\
Target & 96 & 0.0 & 55.2 \\
Food Network & 84 & 0.0 & 39.3 \\
Ulta & 105 & 0.0 & 41.9 \\
\bottomrule
\end{tabular}
\caption{Full-Pro results for the 20 websites with zero detected external-access failures (\%). Together with \autoref{tab:site-blocked-app}, this completes the 42-site ledger.}
\label{tab:site-zero-block-app}
\end{appxtable}

\paragraph{Resource accounting.}
Resource figures are separated from the scientific result tables and are not used to rank controller quality. \autoref{tab:resource-app} reports the archived token and API-cost ledger in a compact form. Additional resources used during experiment stages, hyperparameter tuning, ablations, etc., are not included. 

\begin{appxtable}
\footnotesize
\setlength{\tabcolsep}{1.8pt}
\begin{tabular}{@{}lrrrr@{}}
\toprule
Run & $N$ & Tokens & Total & / rollout \\
\midrule
Full-Pro & 4096 & 268.0M & \$758.75 & \$0.185 \\
Struct teacher & 1024 & 81.98M & \$229.64 & \$0.224 \\
Struct flat & 1024 & 59.49M & \$183.41 & \$0.179 \\
Struct intent & 1024 & 58.79M & \$178.14 & \$0.174 \\
IntentCore & 1024 & 61.65M & \$185.23 & \$0.181 \\
Full-Flash & 4096 & 419.9M & \$196.55 & \$0.048 \\
Flash-60 & 4096 & -- & \$348--355 & \$0.085--0.087 \\
GPT-5.5-Diag & 1024 & $\sim$88.12M & $\sim$\$559.56 & \$0.547 \\
\bottomrule
\end{tabular}
\caption{Proprietary Model API Resource accounting. }
\label{tab:resource-app}
\end{appxtable}

\section{Appendix K: Release and Reproducibility Boundary}
\refstepcounter{appendixref}
\label{app:release-notes}

The release mirrors the paper's unit of analysis: a task contract, a controller intention, a screenshot or observation, one executable action, and verification labels. \autoref{tab:asset-map-app} lists the public artifact classes; \autoref{tab:public-step-schema-app} records the step-row groups needed to reconstruct trajectories and paper metrics.

\begin{appxtable}
\footnotesize
\begin{tabularx}{\linewidth}{@{}>{\raggedright\arraybackslash}p{2.20cm}>{\raggedright\arraybackslash}X@{}}
\toprule
Artifact & Public contents and reconstruction role \\
\midrule
Documentation & overview, loading instructions, and checksums; identifies release entry points \\
Task contracts & 4,096 contracts from 768 audited tasks; reconstructs task, policy, constraints, evidence, and split statistics \\
Step rollouts & screenshot-bearing records for reported full and executable-boundary runs; reconstructs trajectories and metrics \\
Example bundle & 100 task folders with summaries, step records, and screenshots; supports qualitative inspection \\
Evaluation outputs & calibrated contract gates, paired deltas, and stratified summaries \\
Middle training & data/model manifests, offline evaluation, and live replacement summaries \\
Runtime/code & Action AST, guards/compiler, browser runtime, evaluator, and table scripts \\
\bottomrule
\end{tabularx}
\caption{Release asset map. Public artifacts cover the paper's evidence while excluding credentials, private browser state, and restricted third-party material.}
\label{tab:asset-map-app}
\end{appxtable}

\begin{appxtable}
\footnotesize
\begin{tabularx}{\linewidth}{@{}>{\raggedright\arraybackslash}p{1.75cm}>{\raggedright\arraybackslash}X@{}}
\toprule
Row group & Representative fields and reconstruction role \\
\midrule
Task contract & request, hard requirements, persona policy, evidence obligations; identifies the delegated policy \\
Top state & intention, active constraints, missing evidence, blockers, expected observation; explains the requested action \\
Observation & screenshot, URL, title, visible labels, execution status; records what the controller could observe \\
Executable action & action kind, target, text/query, post-guard action; links intention to browser transition \\
Verification & allow-answer, evidence-sufficient, drift, missing obligations; selects continue/recover/answer/fail \\
Outcome & CGS, hard/evidence/answer scores, external-access/runtime fields; reconstructs result tables \\
\bottomrule
\end{tabularx}
\caption{Public step-row schema. Grouping by rollout identifier and sorting by step index reconstructs the trajectory and its contract-gated outcome.}
\label{tab:public-step-schema-app}
\end{appxtable}

Credentials, account state, tokens, browser profiles, and secret-bearing environment files are excluded. The artifact includes consent-cleared human-study aggregates; raw rationales and screenshot material restricted by third-party terms remain outside the public boundary.

\paragraph{Paper-to-code traceability.}
The source release contains a machine-readable implementation map and the same
map as a rendered README.  Every module implementing a novel paper component
starts with a module comment naming the corresponding paper section and its
role: contract/schema construction; persistent top state; middle Action AST;
guards and fallback repair; deterministic execution; stable scheduling;
contract-gated evaluation; human packet construction/aggregation; and
middle-policy export, training, and evaluation.  Generic API, serialization,
and command-line plumbing are identified as infrastructure rather than method
steps.  This convention lets a reader move from each method block in the paper
to its entry point without relying on machine-specific paths.

\section{Appendix J: Worked Contracts and Rollout Cases}
\refstepcounter{appendixref}
\label{app:examples-failures}

\begin{appxtablewide}
\small
\begin{tabularx}{0.78\linewidth}{@{}r p{1.65cm} p{4.40cm} X@{}}
\toprule
Step & Mode & Guarded action / decision & Evidence state \\
\midrule
1 & Query & click continue & public browsing preserved; product evidence missing \\
2 & Act & type constrained query & seek no-drill, wide-window, and odor evidence \\
3 & Query & scroll results & curtain evidence seen; fit and installation remain \\
4 & Inspect & click candidate 1 & no-drill candidate found; width remains \\
5 & Verify & scroll details & inspect size, description, and reviews \\
6 & Verify & scroll details & required product constraints covered \\
7 & Verify & go back & comparison still needs a second candidate \\
8 & Backtrack & click candidate 2 & second no-drill source opened \\
9 & Inspect & scroll details & second product attributes become visible \\
10 & Verify & top-layer decision: answer & evidence gate opens for comparison and caveat \\
\bottomrule
\end{tabularx}
\captionsetup{width=0.78\linewidth}
\caption{Complete action/evidence trace underlying \autoref{fig:rollout-audit}. The top layer delays finalization until two candidates have been inspected; each intermediate record stores the top intention, one guarded action, and the remaining obligations.}
\label{tab:figure-two-trace-app}
\end{appxtablewide}

\begin{figure*}[!tp]
\centering
\includegraphics[width=\linewidth]{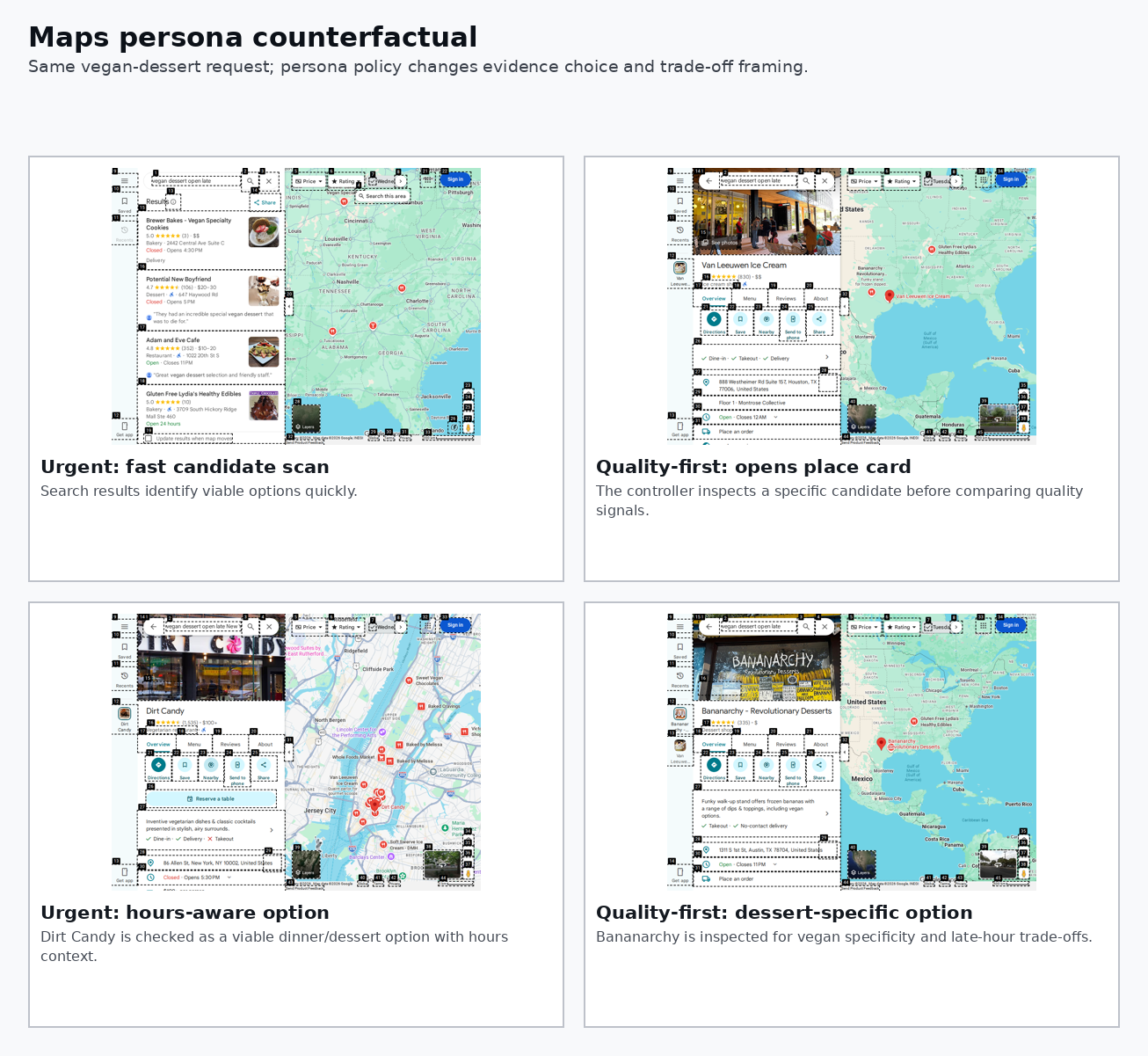}
\caption{Same request, different acceptable routes. The urgent-pragmatist rollout moves from the result list to viable late-hour options; the quality-first rollout opens candidate-specific evidence before recommending. Both pass CGS, but their search depth and evidence choices follow different task-local policies.}
\label{fig:case-google-maps-persona}
\end{figure*}

\begin{figure*}[!tp]
\centering
\includegraphics[width=\linewidth]{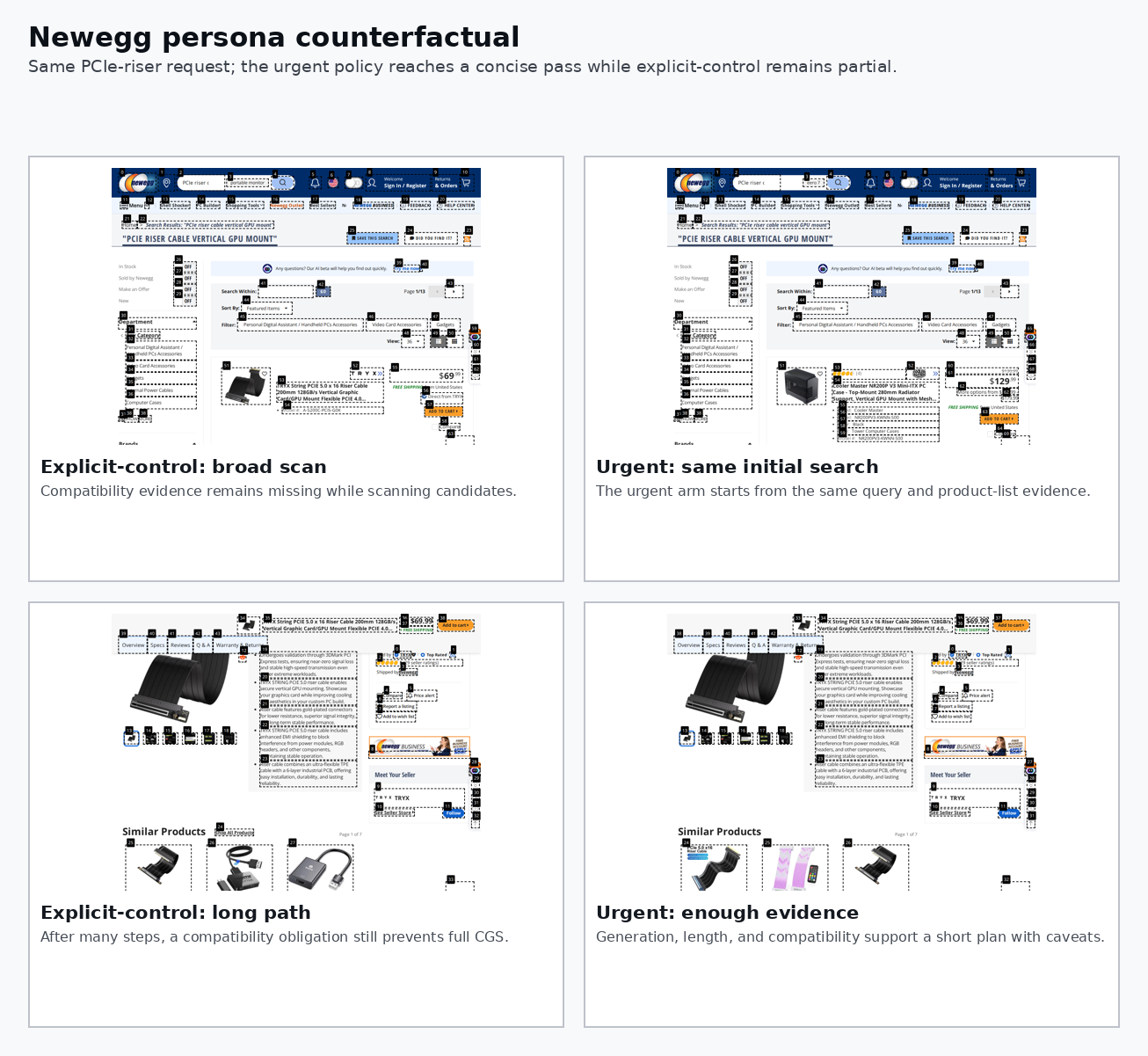}
\caption{Same request, different stopping behavior. The explicit-control rollout remains partial while compatibility evidence is unresolved; the urgent-pragmatist rollout reaches a passing plan after generation, length, and compatibility evidence become visible. The comparison isolates a persona-conditioned evidence threshold under fixed hard constraints.}
\label{fig:case-newegg-persona}
\end{figure*}

\begin{figure*}[!tp]
\centering
\begin{minipage}[t]{0.48\linewidth}
  \centering
  \includegraphics[width=\linewidth]{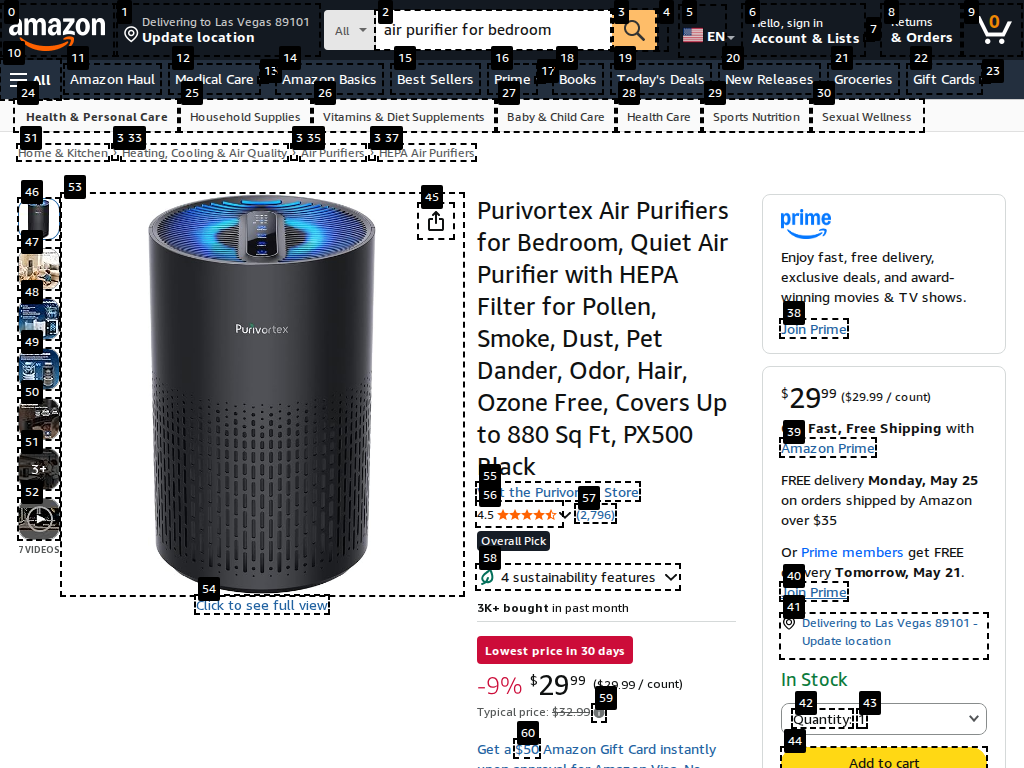}
  \smallskip
  \textbf{A: recommendation grounded on a visible product page}
\end{minipage}\hfill
\begin{minipage}[t]{0.48\linewidth}
  \centering
  \includegraphics[width=\linewidth]{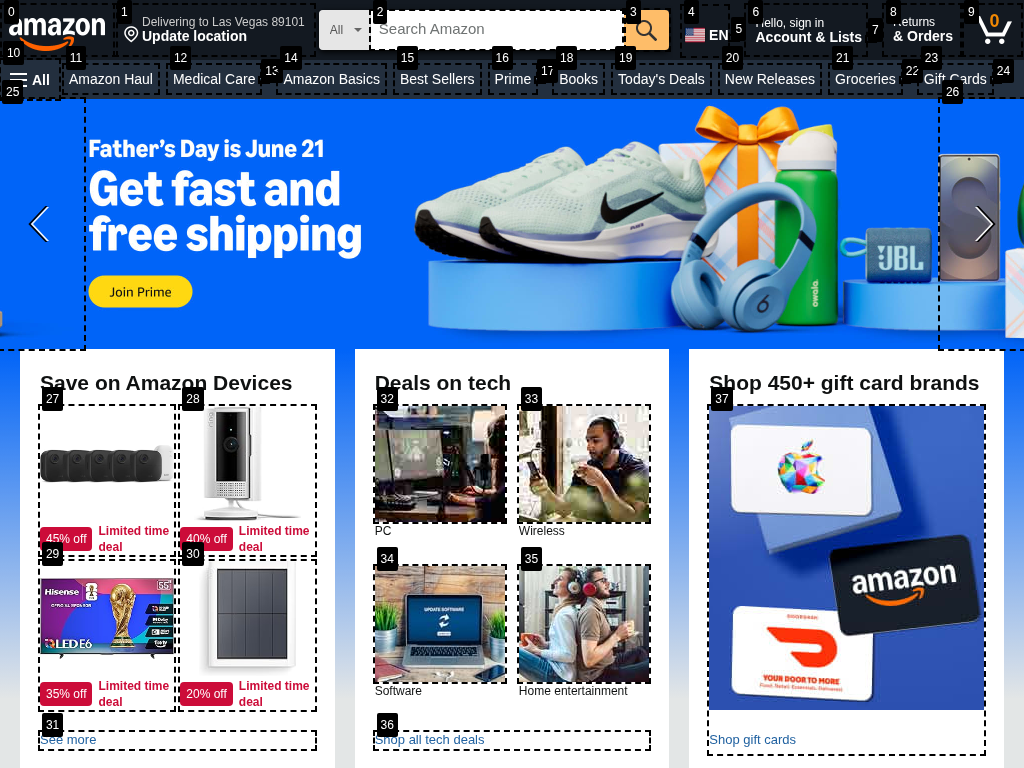}
  \smallskip
  \textbf{B: answer issued from a non-evidentiary browser state}
\end{minipage}
\smallskip

\footnotesize
\setlength{\tabcolsep}{4pt}
\begin{tabular}{@{}l p{5.25cm} p{5.25cm}@{}}
\toprule
 & \textbf{Side A} & \textbf{Side B} \\
\midrule
Policy & safety/privacy guardian & brand loyalist \\
Visible route & click--type--click--scroll--answer & type--search--answer \\
Hard / evidence / answer & 1.00 / 1.00 / 0.90 & 1.00 / 1.00 / 0.90 \\
Persona proxy & 1.00 & 0.50 \\
Terminal browser state & inspected product page & Amazon home page \\
\bottomrule
\end{tabular}
\caption{Two persona-conditioned contracts for the same base request pass the
automatic terminal contract gate but differ in visible process support. Side~B
exposes the permissive boundary of the automatic persona proxy: a CGS pass can still
warrant stepwise human review.}
\label{fig:case-cgs-pass-pcs-low}
\end{figure*}

\begin{figure*}[!tp]
\centering
\includegraphics[width=\linewidth]{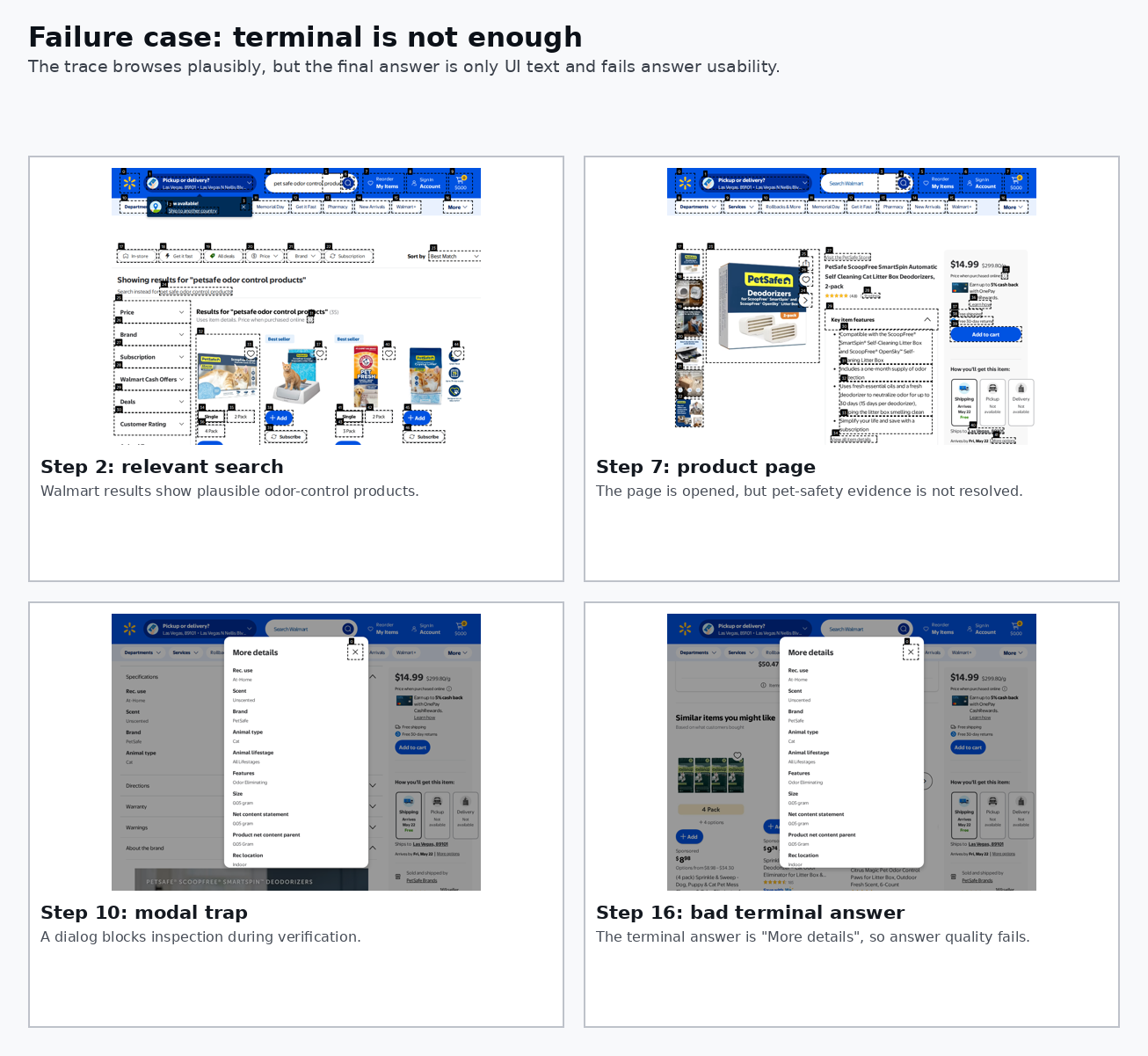}
\caption{Walmart failure case. Relevant browsing is followed by a modal loop and a terminal answer copied from the interface. The example separates locally plausible actions from a usable, evidence-backed completion.}
\label{fig:case-walmart-failure}
\end{figure*}

This appendix connects the formal objects to inspectable examples. \autoref{tab:figure-two-trace-app} exposes the exact trace behind the main rollout figure. \autoref{fig:task-contract-card} shows how a persona-free task becomes counterfactual contracts. \autoref{fig:case-google-maps-persona} and \autoref{fig:case-newegg-persona} show policy-conditioned route and stopping differences; \autoref{fig:case-cgs-pass-pcs-low} and \autoref{fig:case-walmart-failure} show process risks that terminal completion alone would obscure.
\looseness=-1

\paragraph{Exact trace behind \autoref{fig:rollout-audit}.}
The archived Exec-Pro record is task \texttt{rbc4-00001}, a ten-step Amazon comparison of renter-friendly blackout curtains. The main figure displays steps 2, 3, 5, 6, 9, and 10; the full action/evidence sequence is in \autoref{tab:figure-two-trace-app}.

\begin{appxfigure}
\includegraphics[width=0.94\linewidth]{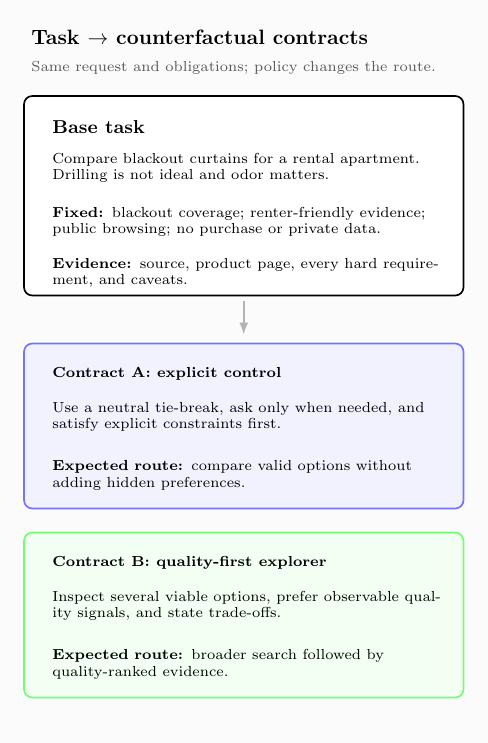}
\caption{A base request becomes two task-local intent contracts. Hard constraints and evidence obligations remain fixed; search breadth, verification, asking, ranking, and stopping may change.}
\label{fig:task-contract-card}
\end{appxfigure}

\paragraph{A CGS false-pass risk.}
Packet \texttt{pcs\_hcp\_0137} contains two persona-conditioned contracts for
the same air-purifier base request. Both receive automatic CGS passes. Side~A
(task \texttt{rbc4-00024}) opens a product page and grounds its recommendation
in visible evidence. Side~B (task \texttt{rbc4-00022}) answers while its final
browser state remains the Amazon home page and introduces Coway and Levoit even
though its brand-loyalist policy forbids assuming an unspecified brand. Its
automatic persona proxy is 0.50, the pass threshold. The case is therefore an
inspectable audit diagnostic, not a human-study observation.

\FloatBarrier

\end{document}